\documentclass[runningheads]{llncs}
\usepackage[T1]{fontenc}
\usepackage{graphicx}
\usepackage{marvosym}
\usepackage{amsmath}
\usepackage{amssymb}
\usepackage{booktabs}
\usepackage{subcaption}
\usepackage{color}
\usepackage{booktabs}
\usepackage[table]{xcolor}
\usepackage{wrapfig} 
\usepackage{multirow}
\begin{document}
\title{Rethinking the Misalignment Problem in Dense Object Detection}
%
%
%
\author{Yang Yang\inst{1,2} \and
Min Li\inst{1,2}({\Letter}) \and Bo Meng\inst{1,3}\and Zihao Huang\inst{1,2}\and\\
Junxing Ren\inst{1,2} \and Degang Sun\inst{1,2}}

\institute{Institute of Information Engineering, Chinese Academy of Sciences, Beijing, China
\and School of Cyber Security, University of Chinese Academy of Sciences, Beijing, China
\and Beijing Institute of Technology, Beijing, China\\
\email{\{yangyang1995, limin, renjunxing, sundegang\}@iie.ac.cn}}
\maketitle              
\begin{abstract}
  Object detection aims to localize and classify the objects in a given image, and these two tasks are sensitive to different object regions. Therefore, some locations predict high-quality bounding boxes but low classification scores, and some locations are quite the opposite. A misalignment exists between the two tasks, and their features are spatially entangled. In order to solve the misalignment problem, we propose a plug-in \textbf{S}pati\textbf{al}-disentangled and \textbf{T}ask-aligned operator (SALT). By predicting two task-aware point sets that are located in each task's sensitive regions, SALT can reassign features from those regions and align them to the corresponding anchor point. Therefore, features for the two tasks are spatially aligned and disentangled. To minimize the difference between the two regression stages, we propose a Self-distillation regression (SDR) loss that can transfer knowledge from the refined regression results to the coarse regression results. On the basis of SALT and SDR loss, we propose SALT-Net, which explicitly exploits task-aligned point-set features for accurate detection results. Extensive experiments on the MS-COCO dataset show that our proposed methods can consistently boost different state-of-the-art dense detectors by $\sim$2 AP. Notably, SALT-Net with Res2Net-101-DCN backbone achieves 53.8 AP on the MS-COCO \emph{test-dev}.

\end{abstract}
 
\section{Introduction}
\label{sec:intro}
\begin{figure}[t]
    \centering
    \centerline{\includegraphics[width=\textwidth]{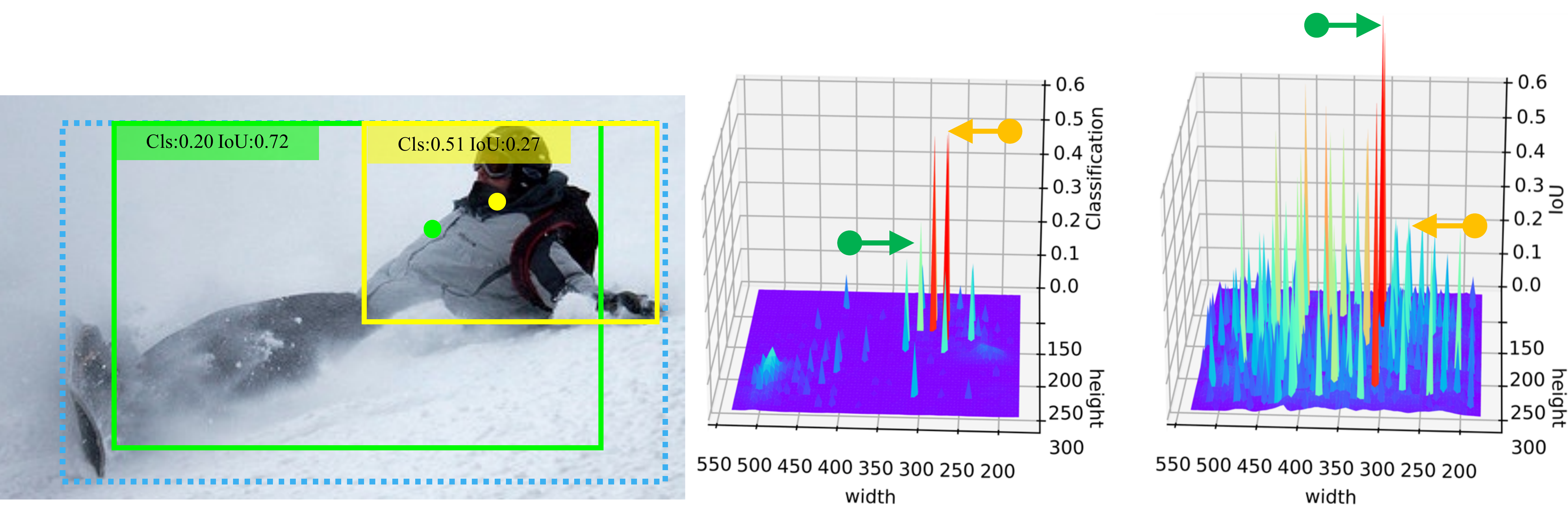}}
    \leftline{\qquad \quad \  (a)Detection Results\qquad \qquad \qquad   (b)Classification\qquad \qquad \ \ \ \  (c)IoU}
    \caption{Illustration of the spatial misalignment of classification and regression. In (a), the blue box denotes the ground truth, and the other boxes are the detection results of ATSS \cite{zhang2020bridging}. The two points are the locations where the detection results are predicted. (b) and (c) are the distributions of classification and IoU scores over all image pixels, and ``IoU'' denotes the intersection over union between the predicted box and the ground truth. } 
    \label{alignment0}
\end{figure}

The main goal of object detection contains two tasks, one is to give the accurate location of the object in an image (i.e., regression), and the other is to predict the category of the object (i.e., classification). During the inference step, the regression and classification results predicted from the same location are paired together as the detection result. Then the NMS algorithm is usually applied to remove redundant detection results by taking the classification scores as the ranking keywords. For the same instance, the detection result with a high classification score will be kept, while others are filtered out. However, the natures of these two tasks are so distinct that they require features from different object locations. As shown in Figure \ref{alignment0}, the classification and regression quality (i.e., IoU) scores from the same location can be quite different. Classification focus on the salient part of the object (e.g., the head of the person), while regression is sensitive to the whole object, especially for its border part.  \emph{Therefore, the prediction distributions of the two tasks are misaligned.} The detection result with a high classification score can have low-quality regression prediction and vice versa.

We model the prediction qualities of the two tasks as two Discrete distributions. Therefore, the goal of solving the misalignment problem is \emph{bridging the gap between these two distributions (i.e., minimizing the distance of their peak positions)}.

\begin{figure}[t]
    \centering
    \begin{minipage}[1]{0.48\linewidth}
		\centerline{\includegraphics[width=\textwidth]{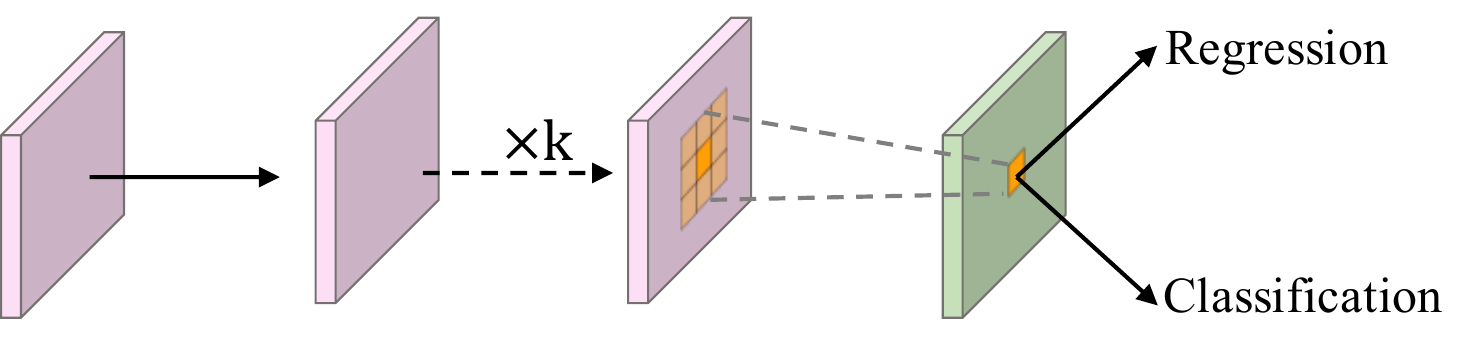}}
      \end{minipage}
      \begin{minipage}{0.48\linewidth}
		\centerline{\includegraphics[width=\textwidth]{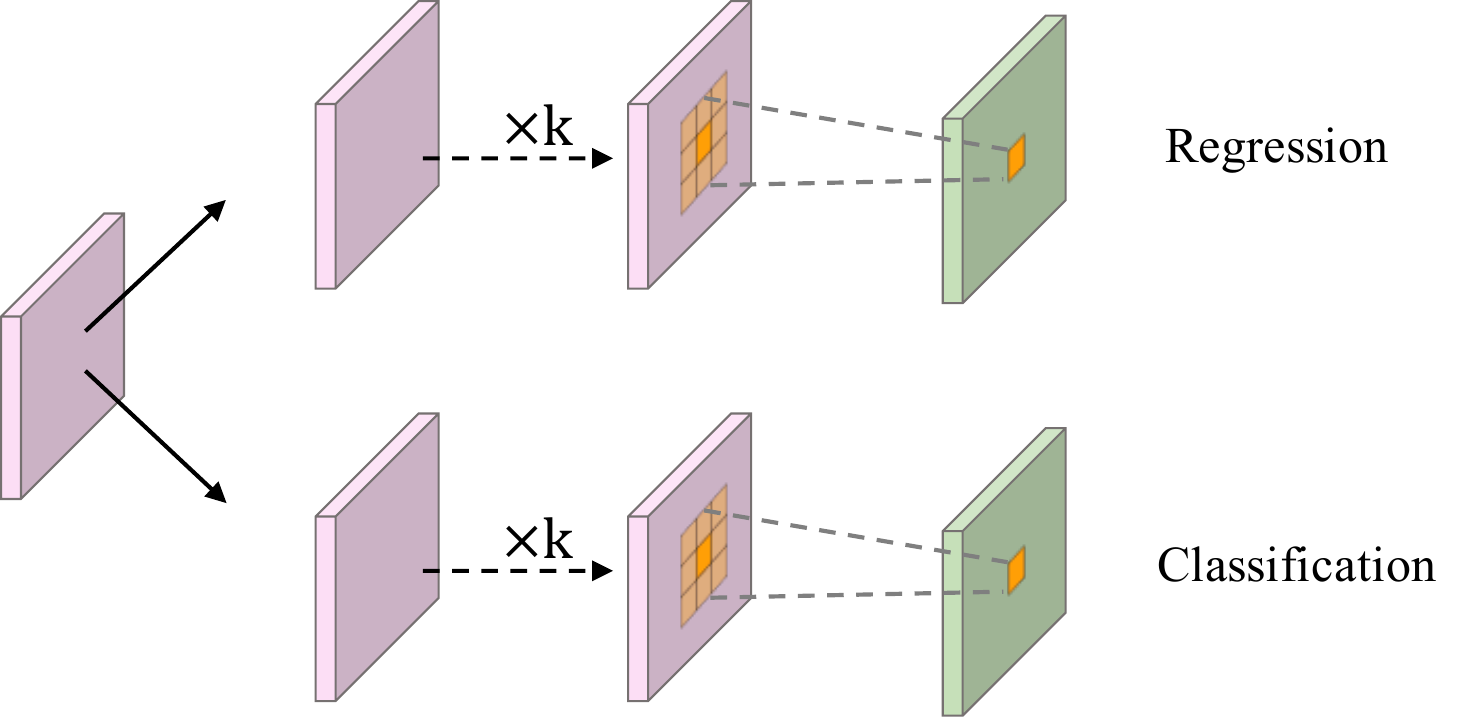}}	
      \end{minipage}
      \centerline{(a)Coupled \qquad \qquad \qquad \qquad \qquad \qquad (b)Decoupled\qquad}
  
    \caption{Illustration of the entangled features in the coupled and decoupled heads.}
    \label{heads}
  \end{figure}

CNN-based dense detectors utilize a coupled or decoupled head to conduct classification and regression. As illustrated in Figure \ref{heads} (a), the coupled head predicts the classification and regression results based on the shared features \cite{yolov3,lin2017feature,redmon2016you}. As a result, the coupled head structure introduces feature conflicts between the two tasks and makes them compromise each other. To solve this problem, the decoupled head structure \cite{wu2020rethinking} is proposed and has been widely adopted in recent years \cite{song2020revisiting,lin2017focal,tian2019fcos}. As shown in Figure \ref{heads} (b), the decoupled head utilizes two parrel sub-networks to perform regression and classification, respectively. This could alleviate the conflict problem by reducing the shared parameters. However, the point features (i.e., the two orange points) that predict the detection result still share the identical receptive field. In conclusion, \emph{both the coupled and decoupled heads predict the classification and regression results from the spatially identical and entangled features}. Considering the difference in their spatial sensitivity, the entangled features inevitably make a location prefer one task over the other one, thereby compounding the misalignment problem.

  In this paper, we propose a plug-in operator to address the misalignment problem: the \textbf{S}pati\textbf{al}-disentangled and \textbf{T}ask-aligned operator (SALT). The first stage of our network is the coarse regression predictions made by a simple Dirac delta decoder \cite{zhang2020bridging}. After that, SALT predicts two sets of spatial-disentangled points to represent each task's sensitive regions, respectively. Then we use bilinear interpolation to reassign features from those regions to the corresponding anchor point.  In the second stage, SALT utilizes spatial-disentangled and task-aligned features to make refined predictions with a General distribution decoder \cite{li2021generalized}. Therefore, a single anchor point can obtain accurate regression and classification predictions simultaneously.  Feature reassignment can bring the peak positions of the two Discrete distributions closer so that SALT can weaken the impact of the misalignment problem.

  In order to minimize the difference between the first and second stage predictions, we also propose a novel self-distillation regression (SDR) loss, making the coarse predictions learn from the refined predictions. As a result, the final performance got improved without any extra inference cost.

  \begin{enumerate}
    \item We propose an operator that can generate spatial-disentangled and task-aligned features for regression and classification, respectively. 
    \item The proposed operator can be easily plugged into most dense object detectors and bring a considerable improvement of $\sim$2 AP.
    \item Our proposed SDR loss can also boost the overall performance in an inference cost-free fashion.
    \item Without bells and whistles, our best single-scale model (Res2Net-101-DCN) yields 51.4 AP on the COCO \emph{test-dev} set, which is very competitive results among dense object detectors.
  \end{enumerate}

  \section{Related Work}
  \subsubsection{Misalignment:}
  Dense detectors, such as IoU-aware \cite{wu2020iou}, FCOS \cite{tian2019fcos} and PAA \cite{kim2020probabilistic} apply an extra branch to predict the regression confidence and combine it with the classification confidence as the detection score. Different from previous methods, GFL \cite{li2020generalized} and VFNet \cite{zhang2021varifocalnet} propose a joint representation format by merging the regression confidence and classification result to eliminate the inconsistency between training and inference. TOOD \cite{feng2021tood} proposes a prediction alignment method that predicts the offset between each location and the best anchor and then readjusts the prediction results. Guided Anchoring \cite{wang2019region}, RefineDet \cite{zhang2018single}, and SRN \cite{chi2019selective} learn an offset field for the preset anchor and then utilize a feature adaption module to extract features from the refined anchors. RepPoints \cite{yang2019reppoints} and VFNet \cite{zhang2021varifocalnet} utilize the deformable convolution \cite{dai2017deformable} to extract accurate point feature. However, all the aforementioned methods extract features for regression and classification from the same locations, without considering their spatial preference. That is, the features for these tasks are spatially entangled, which leads to inferior performance. 

  \subsubsection{Self-distillation:}
  Model distillation \cite{hinton2015distilling} usually refers to transferring knowledge from a pre-trained heavy teacher network to a compact student network. DML \cite{zhang2018deep} provides a new paradigm that a pre-trained teacher is no longer needed and all the student counterparts are trained simultaneously in a cooperative peer-teaching manner. Following this paradigm, many self-distillation approaches \cite{zhang2019your,yao2020knowledge,ji2021refine,li2021student} are proposed for classification knowledge transfer learning. However, transferring regression knowledge of object detection has been proven to be difficult \cite{kang2021instance,wang2019distilling}, as different locations of an image have different contributions to the regression task. LGD \cite{zhang2021lgd} is the only self-distillation approach for general object detection, which proposes an intra-object knowledge mapper that generates a better feature pyramid and then performs distillation with feature imitation. This approach provides performance gains but also introduces too many auxiliary layers.
  
  \begin{figure*}[t]
    \centering
    \includegraphics[width=0.99\textwidth]{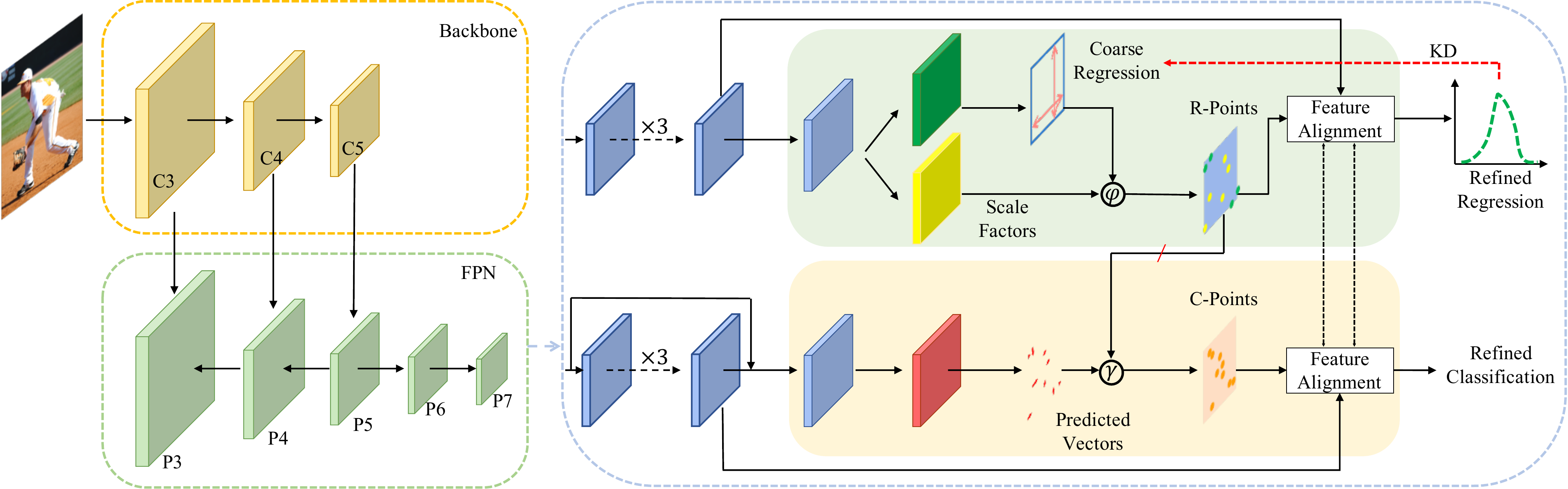} 
    \caption{Architecture of SALT-Net. Our proposed architecture consists of a backbone, an FPN (P3-P7), and two subnetworks for classification and regression, respectively.
      ``$\varphi$" and ``$\gamma$" denote Equations (\ref{ps}) and (\ref{disentangle}), respectively. ``\textcolor[rgb]{1,0,0}{$/$}" denotes the gradient flow detachment. ``KD" denotes our proposed self-distillation approach. ``R-points" and ``C-Points" denote the regression-aware and classification-aware points, respectively.}
    \label{network} 
    \end{figure*} 
  
  \section{Proposed Approach}
  
  In this section, we first detail the proposed operator SALT. Then we introduce our self-distillation approach that enables the first-stage decoder to learn from the second-stage decoder. Finally, we introduce the loss function of SALT-Net. The network architecture (Figure \ref{network}) and inference details can be found in the supplementary material.

  \begin{wrapfigure}{r}{0.49\textwidth}
    \vspace{-36pt}
    \begin{center}
        \setlength{\fboxrule}{0pt}
        \setlength{\abovecaptionskip}{3pt}
        \fbox{\includegraphics[width=0.47\columnwidth]{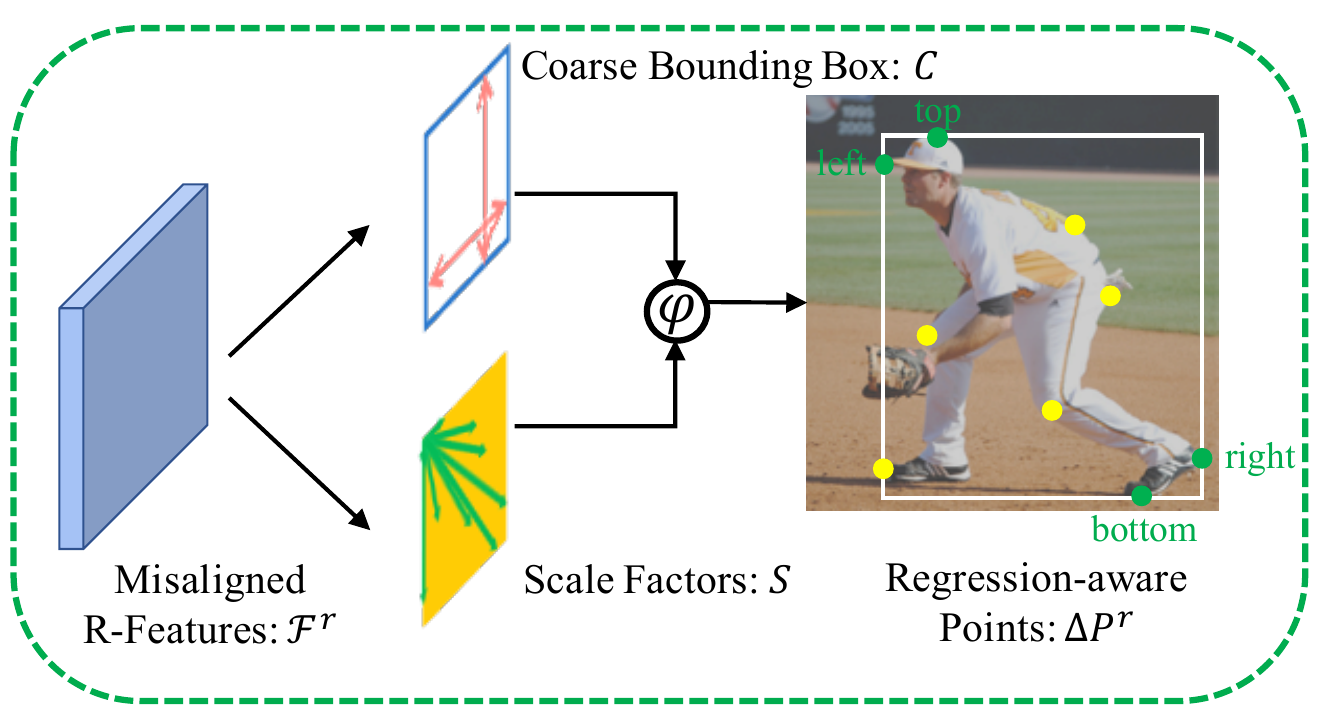}}
    \end{center}	
    \vspace{-12pt}
    \caption{Illustration of the regression-aware points. ``$\varphi$" and the white box denotes Equation (\ref{ps}) and the coarse bounding box prediction, respectively.}
  \label{re}  
    \vspace{-8pt}
\end{wrapfigure}
  
\subsection{SALT: Regression-aware Points}
As shown in Figure \ref{re}, given the misaligned regression features $\mathcal {F}^r$ from the last layer of the regression tower (i.e., the 3$\times$ convolutions shown in Figure \ref{network}), SALT first predicts the coarse bounding box $\mathcal {C}$ with the Dirac delta decoder, as in \cite{tian2019fcos,zhang2020bridging}. The coarse bounding box is represented by the top-left corner and its width and height (i.e., $(xmin, ymin, w, h)$). 

Then SALT predicts the scale Factors $\mathcal {S}$ that measures the normalized distances between the top-left corner of the coarse bounding box and the regression-sensitive regions (i.e., regression-aware points $\Delta \mathcal P^r$). Scale factors $\mathcal S$ and the coarse bounding box $\mathcal C$ are obtained by only two convolution layers, i.e.:  
\begin{equation}    
\left\{\begin{array}{c} \mathcal C= \delta({conv}_{c}(\mathcal {F}^r))  
\\ \mathcal S= \sigma({conv}_{s} (\mathcal {F}^r))\end{array}\right.
\label{bs} 
\end{equation}
where $\sigma$ and $\delta$ are Sigmoid and ReLU, respectively. $\mathcal C \in \mathbb{R}^{H\times W\times 4}$, $\mathcal S \in \mathbb{R}^{H\times W\times 2N-4}$ and $N$ is the number of the regression-aware points. Then the location of $i$-th regression-aware point $\Delta p_i$ can be obtained with Equation (\ref{ps}):
  \begin{equation}
  \left\{\begin{array}{c} x_{i} =x_{\min }+w * {\mathcal S}_{i x}  \\ y_{i} =y_{\min }+h * {\mathcal S}_{i y}\end{array}\right. 
      \label{ps}
  \end{equation}
  where $(xmin,ymin)$ is the location of the top-left corner of the coarse bounding box $\mathcal C$, and $({\mathcal S}_{i x},{\mathcal S}_{i y})$ are scale factors that measure the normalized distance between the $i$-th point and top-left corner. Therefore, the location of the regression-aware points $\Delta \mathcal P^r$ can be represented by:
  \begin{equation}
      \Delta \mathcal P^r={\{\Delta p_i^r\}^N_{i=1}}
  \end{equation}

Note that all coordinates are represented by taking the location that predicts the detection result as the coordinate origin. Therefore, the coordinates mentioned in this section are relative locations, not absolute coordinates. 

The total number of the regression-aware points is $N$, the channels of scale factors $\mathcal S$ and the regression-aware points $\Delta \mathcal P^r$ are  $2N-4$ and $2 N$, respectively. The reason for this inconstancy is that we want to ensure that the sampled regression-aware points contain the four extreme points (i.e., left-most, right-most, top-most, bottom-most), which encode the location of the object. On this account, four points are sampled on the four bounds of the coarse bounding box (i.e., the green points in Figure \ref{re}), respectively. As the location of the bounding box has been predicted, four axial coordinates of the extreme points are preset and do not need to be learned (i.e., $xmin, ymin, xmin+w, ymin+h$).  
  
    \subsection{SALT: Classification-aware Points}
    \begin{wrapfigure}{r}{0.49\textwidth}
        \vspace{-35pt}
        \begin{center}
            \setlength{\fboxrule}{0pt}
            \setlength{\abovecaptionskip}{3pt}
            \fbox{\includegraphics[width=0.47\columnwidth]{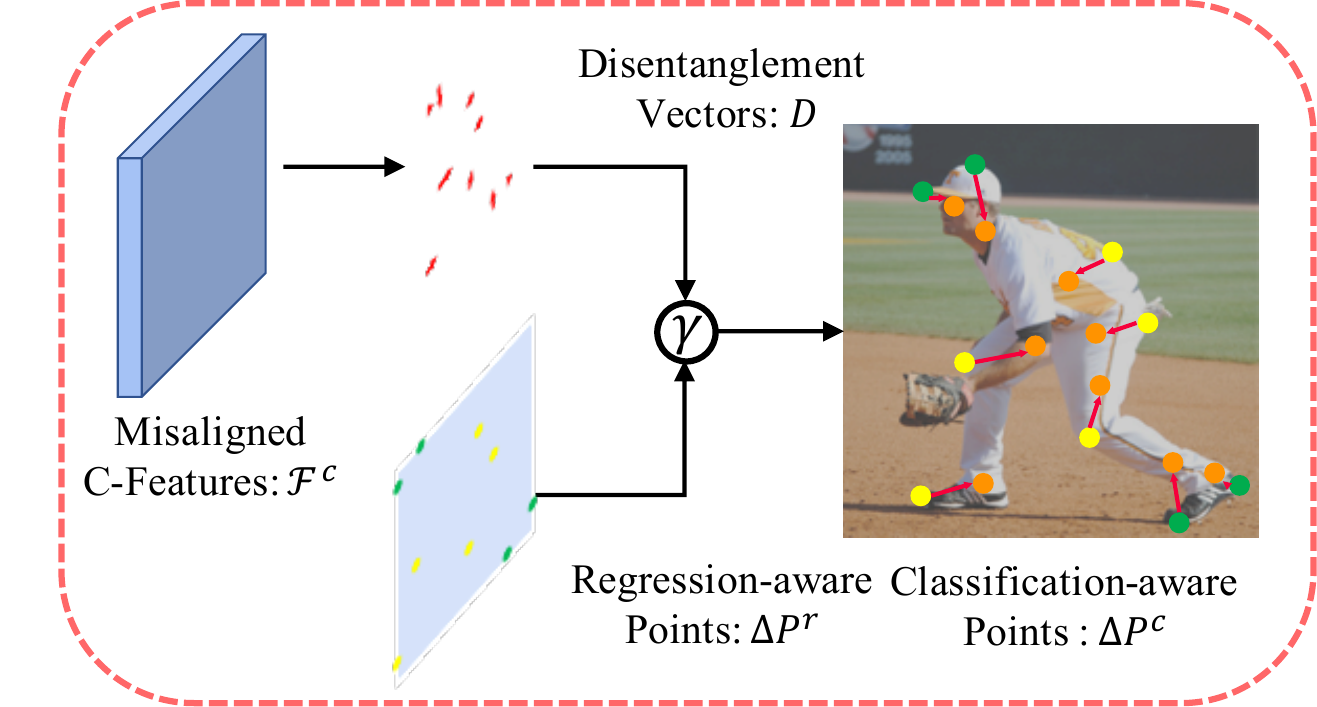}}
        \end{center}	
        \vspace{-12pt}
        \caption{Illustration of the spatial disentanglement method. ``$\gamma$" denotes Equation (\ref{disentangle}).}
      \label{alignment}  
        \vspace{-8pt}
    \end{wrapfigure}
Regression and classification are sensitive to different areas of the object. For this reason, extracting features from the regression-interested-locations hinders the detection performance. Therefore, SALT contains a spatial disentangle module to guide the classification branch to generate a set of classification-aware points. As shown in Figure \ref{alignment}, the regression-aware points act as the shape hypothesis of the object to be classified. In other words, we take the regression-aware points as a point-set anchor for predicting the classification-aware points.

Similar to the scale factors, this module also consists of only one convolution layer. As shown in Figure \ref{alignment}, given the feature map $\mathcal {F}^c$ from the last layer of the classification tower, the disentanglement vectors $\mathcal D$ are obtained by:
    \begin{equation}
        \mathcal D= \delta({conv}_{d} (\mathcal {F}^c))
        \label{dis}
    \end{equation}
    
    With the regression-aware points $\Delta \mathcal P^r$ taken as the point-set anchor, we propose two functions to generate the classification-aware points, as illustrated by Equation (\ref{disentangle}) and (\ref{disentangleplus}). We choose Equation (\ref{disentangle}) as the final prediction strategy. Details and analysis can be found in Sec. \ref{sd}. 
    
    \begin{equation}
      \Delta \mathcal P^c=e^{\mathcal D} \cdot \Delta \mathcal P^r
      \label{disentangle}
    \end{equation}
    \begin{equation}
      \Delta \mathcal P^c={\mathcal D} + \Delta \mathcal P^r
      \label{disentangleplus}
    \end{equation}
  
    To make sure the learning process of classification and regression are independent of each other. The gradient flow of the regression-aware points $\Delta \mathcal P^r$ is detached from the classification branch. $\Delta \mathcal P^r$ only serves as the prior knowledge in this module. Therefore, the supervision of the classification task does not affect the learning of regression-aware points.    
  
  \subsection{SALT: Feature Alignment}

\begin{wrapfigure}{r}{0.49\textwidth}
\vspace{-35pt}
\begin{center}
    \setlength{\fboxrule}{0pt}
    \setlength{\abovecaptionskip}{3pt}
    \fbox{\includegraphics[width=0.47\columnwidth]{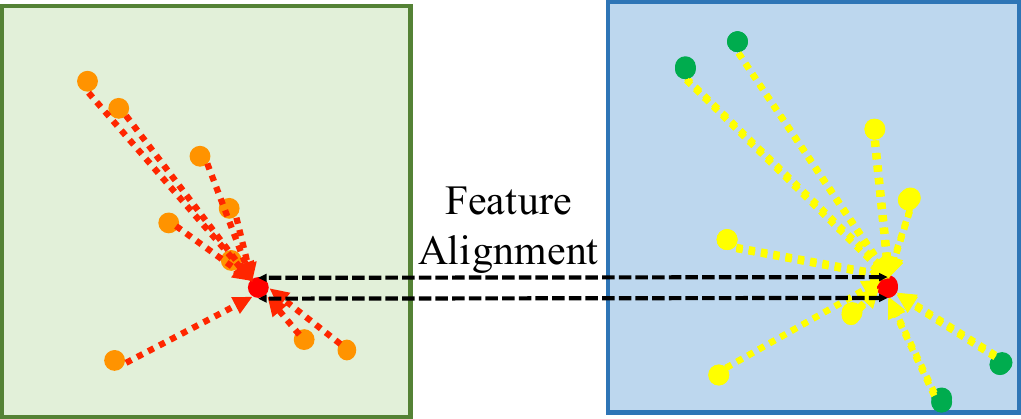}}
\end{center}	
\vspace{0pt}
\caption{Illustration of the feature alignment method. The red point denotes the location that predicts the detection result.}
\label{fa}  
\vspace{-30pt}
\end{wrapfigure}

The regression-aware and classification-aware points are located in each task's sensitive regions, and they are spatially misaligned. Therefore, we aggregate features from those regions to the same anchor point (shown in Figure \ref{fa}). Given the learned point set $\Delta \mathcal P={\{\Delta p_i\}^N_{i=1}}$, we use the bilinear interpolation to make it differentiable. Let ${\{p_i\}^{N}_{i=1}}$ be the sampling window of a regular grid, where $N$ is the number of points. The new irregular sampling locations can be represented by Equation (\ref{newpoints}), and the bilinear interpolation is formulated as Equation (\ref{bilinear}),

  \begin{equation}
    \begin{aligned}
        &&\hat{\mathcal P}=\left\{p+p_i+\Delta {p}_{i} \mid i=1, \ldots, N\right\}
    \label{newpoints}
    \end{aligned}
    \end{equation}
  
    \begin{equation}
      {\hat{\mathcal {F}}}(p)=\sum_{\hat{p}}{G}(\hat{p},p)\cdot{\mathcal {F}}(\hat{p})
      \label{bilinear} 
      \end{equation}
where ${\mathcal {F}}(\cdot)$  and $\hat{\mathcal {F}}(\cdot)$ are the input and output feature maps, and $G(\cdot,\cdot)$ is the bilinear interpolation kernel. $\hat{p} \in \hat{\mathcal P}$, and $p$ is the location that predicts the detection result.
  
The aligned task features are extracted from the locations of the task-aware points, and then they are used for classification and regression refinement. Different from the first stage, the second regression stage utilizes the General distribution decoder \cite{li2020generalized} that outputs the discrete representation of the bounding box.
  \subsection{Self-distillation}
 \begin{wrapfigure}{r}{0.43\textwidth}
        \vspace{-55pt}
        \begin{center}
            \setlength{\fboxrule}{0pt}
            \setlength{\abovecaptionskip}{3pt}
            \fbox{\includegraphics[width=0.4\columnwidth]{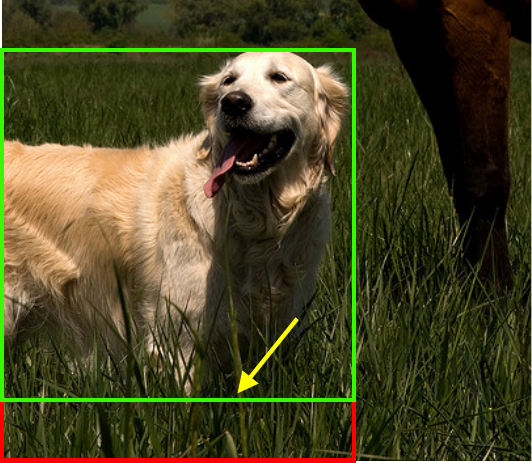}}
        \end{center}  
        \caption{The red and green boxes are the ground truth and the bounding box prediction made by our stage-two decoder, respectively.}
      \label{amb}  
        \vspace{-25pt}
      \end{wrapfigure}
      As Figure \ref{amb} shows, the bottom boundary of the handcrafted annotation is inaccurate and ambiguous, which can misguide and hurt the training process. However, the network's prediction results sometimes provide better and clearer regression targets that are easier for the network to learn. For this reason, we propose a self-distillation regression loss (SDR) that could transfer regression knowledge from the refined predictions to the coarse ones. 

      As Equation \ref{sdr} shows, $R_1$, $R_2$, and $y$ are the output of the stage-one, stage-two decoders, and the classification score. $IoU_1$ and $IoU_2$ denote the Intersection over Union between the ground truth label and the corresponding regression results, and $GIoU$ is the Generalized Intersection over Union as in \cite{rezatofighi2019generalized}. As the stage-two decoder is fed with better features, we take its outputs as the regression upper bound of the stage-one decoder. That is, we utilize the integral results from the Discrete distribution decoder as the soft target for the Dirac delta decoder. Here, $y \cdot IoU_2$ denotes the confidence score of the refined regression result, and its gradient is detached. Thus, SDR loss pays more attention to the high-confidence prediction results. Notably, SDR loss only penalizes the network when the predictions of the stage-two decoder are better than the stage-one decoder (i.e., $IoU_2>IoU_1$). Our proposed SDR loss enables the coarse predictions to learn from the refined results and bridges the gap between them. Better stage-one predictions lead to better stage-two predictions and promote the training process into positive circulation.
    \begin{equation}
        \begin{aligned}
            \mathcal{L}_{SDR}(R_1,R_2,y)=\left\{\begin{array}{l}y \cdot IoU_2 \cdot (1-GIoU(R_1,R_2)),   \ \text{ if } IoU_2>IoU_1 \\
                0  \qquad  \qquad  \qquad  \qquad  \qquad\qquad \ \ \ \text { otherwise }
                \end{array}\right.
                \label{sdr}
        \end{aligned}
    \end{equation}

    \subsection{Loss Function}
    The proposed SALT-Net is optimized in an end-to-end fashion, and both the coarse and the refined detection stages utilize ATSS \cite{zhang2020bridging} as the positive and negative targets assignment strategy. The training loss of SALT-Net is defined as follows:
  \begin{equation}
      \begin{aligned}
      \mathcal{L}&=\frac{1}{N_{\text {pos }}} \sum_z \lambda_{0} \mathcal{L}_{\mathcal{Q}}\\
      &+\frac{1}{N_{\text {pos }}} \sum_z \textbf{1}_{\left\{c_{z}^{*}>0\right\}}\left(\lambda_{1} \mathcal{L}_{\mathcal{R}_1}+\lambda_{2} \mathcal{L}_{\mathcal{R}_2}+\lambda_{3} \mathcal{L}_{\mathcal{D}}+\lambda_{4} \mathcal{L}_{\mathcal{SDR}}\right)
      \end{aligned}
  \end{equation}
  where $\mathcal{L}_{\mathcal{Q}}$ is the Quality Focal loss \cite{li2020generalized} for the classification task. $\mathcal{L}_{\mathcal{R}_1}$ and $\mathcal{L}_{\mathcal{R}_2}$ are both $GIoU$ loss \cite{rezatofighi2019generalized}, one for the coarse bounding box prediction and the other for the refined regression result. $\mathcal{L}_{\mathcal{D}}$ is the Distribution Focal Loss \cite{li2020generalized} for optimizing the general distribution representation of the bounding box, and $\mathcal{L}_{\mathcal{SDR}}$ is the proposed self-distillation loss. $\lambda_{0}\sim\lambda_{4}$ are the hyperparameters used to balance different losses, and they are set as $1, 1, 2,0.5$, and $1$, respectively. $N_{\text {pos }}$ denotes the number of selected positive samples, and $z$ denotes all the locations on the pyramid feature maps. $\textbf{1}_{\left\{c_{z}^{*}>0\right\}}$ is the indicator function, being $1$ if $c_{z}^{*}>0$ and $0$ otherwise.

    \section{Experiments}
    Figure \ref{network} presents the network of our proposed SALT-Net. We take state-of-the-art dense detectors ATSS \cite{zhang2020bridging} and GFLv2 \cite{li2021generalized} as our baseline, and they serve as the stage-one and stage-two decoders, respectively. Our SALT-Net is evaluated on the challenging MS-COCO benchmark \cite{lin2014microsoft}. Following the common practice, we use the COCO train2017 split (115K images) as the training set and the COCO val2017 split (5K images) for the ablation study. To compare with state-of-the-art detectors, we report the COCO AP on the \emph{test-dev} split (20K images) by uploading the detection result to the MS-COCO server.
    
    \subsection{Performance of SALT's component parts}
    \begin{table*}[t]
        \caption{Ablation study of SALT on the COCO val2017 split. $S_1$ and $S_2$ denote the stage-one and stage-two regression results. ``R-Points"  and ``C-Points" denote the regression-aware and Classification-aware points, respectively. ``P-anchor" denotes utilizing the regression-aware points as the point-set anchor for generating the Classification-aware points. ``skip" denotes the skip connection of the classification tower, as shown in Figure \ref{network}.}
        \centering
        {
        {\begin{tabular}{l|cccc|c|cc|ccc} 
        \toprule
        Method&R-Points&C-Points&P-anchor&skip&AP&$\mathrm{AP}_{50}$&$\mathrm{AP}_{75}$&$\mathrm{AP}_{S}$&$\mathrm{AP}_{M}$&$\mathrm{AP}_{L}$\\ 
        \hline
        \hline
        \textbf{baseline}\\ 
        $S_1$\cite{zhang2020bridging}&&&&&39.9&58.5&43.0&22.4&43.9&52.7     \\
        $S_2$\cite{li2021generalized}&&&&&40.9&58.3&44.4&23.9&44.7&53.5     \\
        \hline 
        $S_2$&\checkmark&&&&41.3&58.7&44.9&23.3&45.0&54.2     \\
        $S_2$&\checkmark&\checkmark&&&41.6&59.1&45.6&23.7&45.4&54.7     \\   
        $S_2$&\checkmark&\checkmark&\checkmark&&42.1&59.6&45.6&24.8&45.4&55.5    \\       
        $S_2$&\checkmark&\checkmark&\checkmark&\checkmark&\textbf{42.5}&\textbf{60.1}&\textbf{46.2}&\textbf{25.1}&\textbf{45.9}&\textbf{56.4}\\
        \hline
        $S_1$&\checkmark&\checkmark&\checkmark&\checkmark&41.3&58.6&44.9&23.2&45.1&54.2    \\
        
        \bottomrule
        \end{tabular}}}
        
        \label{Individual}
        \end{table*}

\begin{figure*}[t]
    \centering
    {\begin{subfigure}{0.99\textwidth}
        \includegraphics[width=0.99\textwidth]{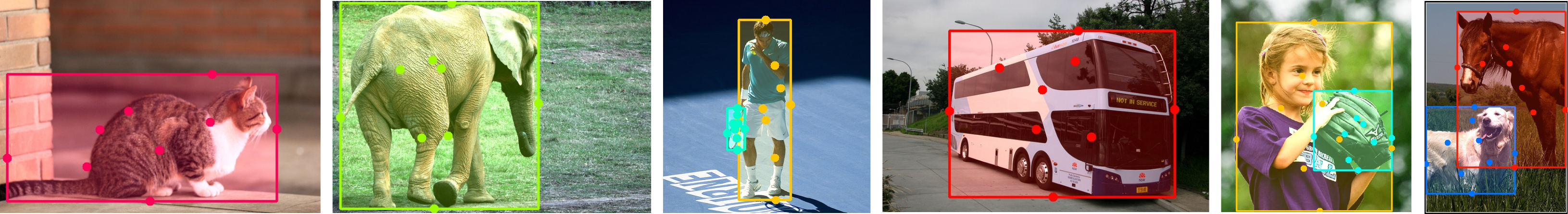}
    \end{subfigure}}
    
    {\begin{subfigure}{0.99\textwidth}
        \includegraphics[width=0.99\textwidth]{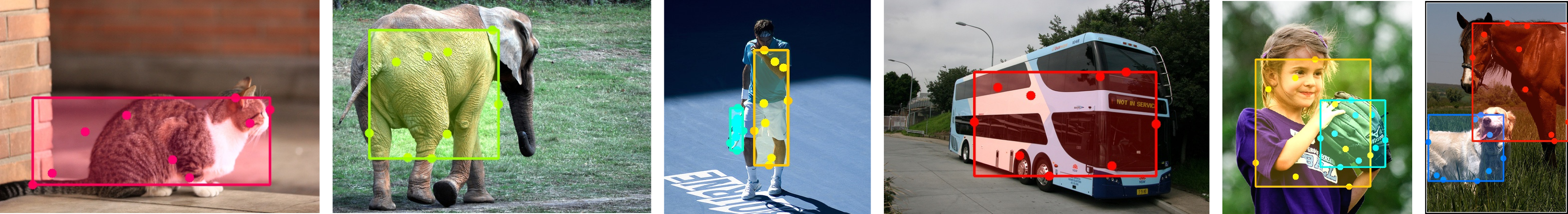}
    \end{subfigure}}
    
    \caption{Visualization of the regression-aware (upper row) and classiﬁcation-aware (lower row) points. Different task-aware points are located on the different areas of the object, and their sensitive regions (i.e., the bounding boxes) are distinct.}
    \label{imgdemo}
    \end{figure*}
To validate the effectiveness of different component parts of our proposed operator SALT, we gradually add the proposed modules to the baseline. As shown in Table \ref{Individual}, the second and third rows are the baseline performances of the stage-one and stage-two decoders, respectively. Note that the stage-one decoder utilizes joint representation of IoU and classification scores instead of its original centernesss branch, as in \cite{li2020generalized}. The baseline performances of the two stages are 39.9 AP and 40.9 AP, respectively. 
  
As presented in the fourth row, the first experiment investigates the effect of implementing the regression-aware points. Therefore, SALT only predicts the scale factors $\mathcal S$ for generating the regression-aware points. Both subnetworks utilize aligned features from the locations of the regression-aware points for the refined detection results. The AP is improved to 41.3, which indicates that the aligned features do improve the detection accuracy, even though features for the two tasks are still spatially entangled.

As shown in the fifth row, to test the effect of spatial disentanglement, SALT predicts the disentanglement vectors $\mathcal D$ for generating the classification-aware points. Note that these points are learned without the regression-aware points acting as the point-set anchor (i.e., $\Delta \mathcal P^c={\mathcal D}$), yet the AP is still boosted to 41.6. 
  These classification-aware points are located in different regions from the regression-aware points, and higher accuracy is obtained (41.6 {vs.} 41.3). Therefore, spatial disentanglement does raise the detection performance by eliminating their spatial feature conflicts. 
  \begin{table}[t]
    \centering
    \caption{Spatial disentanglement strategies. ``$exp$" and ``$+$" denotes utilizing Equation (\ref{disentangle}) and (\ref{disentangleplus}) to generate the classification-aware points, respectively. }
    
    {\begin{tabular}{c|c|cc|ccc} 
    \toprule
    Method&$\mathrm{AP}$&$\mathrm{AP}_{50}$&$\mathrm{AP}_{75}$&$\mathrm{AP}_{S}$&$\mathrm{AP}_{M}$&$\mathrm{AP}_{L}$\\ 
    \hline
    \hline
    w/ $+$&42.3&60.1&46.1&24.7&46.0&56.1     \\
    w/ $exp$&\textbf{42.5}&\textbf{60.1}&\textbf{46.2}&\textbf{25.1}&\textbf{45.9}&\textbf{56.4}     \\
    \bottomrule
    \end{tabular}}
 
    \label{Disentanglement}
    \end{table}

    \begin{table}[t]
      \caption{Performance of implementing our proposed approach in popular dense detectors. }
      \centering
      {\begin{tabular}{l|l|ll} 
      \toprule
      Method&$\mathrm{AP}$&$\mathrm{AP}_{50}$&$\mathrm{AP}_{75}$\\ 
      \hline
      \hline
      FCOS&38.6&57.2&41.7 \\
      \rowcolor{blue!5} SALT-FCOS&\textbf{40.9} \textcolor[rgb]{1,0,0}{(+2.3)}&59.3&44.2 \\
      \hline
      RepPoints w/ GridF&37.4&58.9&39.7 \\
      \rowcolor{blue!5} SALT-RepPoints&\textbf{39.0} \textcolor[rgb]{1,0,0}{(+1.6)}&60.5&41.6 \\ 
      \bottomrule
      \end{tabular}}
  
  \label{plug} 
  \end{table}
The sixth row shows the performance when taking the regression-aware points as the point-set anchor for generating the classification-aware points. It can be observed that a notable performance gain is achieved (i.e., 0.5 AP improvement). That thereby proves the effectiveness of utilizing the regression-aware points as the shape hypothesis and the importance of task disentanglement. Figure \ref{imgdemo} is the visualization of task-aware points and their sensitive regions. This figure indicates that classification and regression are sensitive to different locations of the object, which also gives the interpretability of spatial disentanglement.

As shown in the seventh row, the long-range skip connection (i.e., the residual connection on the classification tower) can also bring a considerable performance boost and gain 0.4 AP. Note that the overall performance has been improved by 1.6 AP and 2.9 AP$_L$ compared with the strong baseline. More details about the skip connection experiments can be found in the supplementary material. Finally, the last row indicates that the coarse regression results with the refined classification results can also improve the baseline performance by 1.4 AP.
 
  \subsection{The Selection of Spatial Disentanglement Strategies}\label{sd}
    We propose two disentanglement functions to generate the classification-aware points, as illustrated in Equation (\ref{disentangle}) and (\ref{disentangleplus}). In Equation (\ref{disentangle}), the disentanglement vector set ${\mathcal D}$ is taken as the exponent, whereas ${\mathcal D}$ is directly aggregated with $\Delta \mathcal P^r$ in Equation (\ref{disentangleplus}).  As illustrated in Table \ref{Disentanglement}, the ``$exp$" strategy performs better than that of ``$+$." The reason is that predicting log-space transforms (i.e., $\mathcal D=\ln{\frac{\Delta \mathcal P^c}{\Delta \mathcal P^r}}$), instead of directly predicting the distance (i.e., $\mathcal D=\Delta \mathcal P^c-\Delta \mathcal P^r$),  prevents unstable gradients during training. Therefore, it is easier to be learned.

  \subsection{Generality of SALT}  
Our proposed SALT can act as a plug-in operator for dense detectors. Therefore, we plug SALT into popular detectors \cite{tian2019fcos} and \cite{yang2019reppoints}, to validate its generality. As shown in Table \ref{plug}, the performance gain is 2.3 AP on FCOS, which is a considerable improvement. Compared with  RepPoints, our SALT-RepPoints performs better than it and gains 1.6 AP. One can see that SALT can significantly improve the accuracy of different detectors, which demonstrates its generality.

\subsection{Self-distillation Regression Loss}
\begin{table}[t]
    \caption{The effect of SDR loss}
    \centering
    {\begin{tabular}{l|c|l|ll|lll} 
    \toprule
    Method&SDR&$\mathrm{AP}$&$\mathrm{AP}_{50}$&$\mathrm{AP}_{75}$&$\mathrm{AP}_{S}$&$\mathrm{AP}_{M}$&$\mathrm{AP}_{L}$\\ 
    \hline
    \hline
    $S_1$&&41.3&58.6&44.9&23.2&45.1&54.2    \\
    $S_2$&&42.5&60.1&46.2&25.1&45.9&56.4\\
    \hline

    \rowcolor{blue!5}$S_1$&\checkmark&\textbf{42.1}\textcolor[rgb]{1,0,0}{(+0.8)}&60.5&45.9&24.8\textcolor[rgb]{1,0,0}{(+1.6)}&45.7&54.9    \\
    \rowcolor{blue!5}$S_2$&\checkmark&\textbf{42.8}\textcolor[rgb]{1,0,0}{(+0.3)}&60.6&46.7&25.1&46.4&56.0\\
    
    \bottomrule
    \end{tabular}}

\label{table:sdr} 
\end{table}

The baseline for this ablation study is the best model of Table \ref{Individual}. Here, both stages utilize the refined classification scores as the NMS ranking keywords. As Table \ref{table:sdr} shows, after applying the SDR loss to the SALT-Net, the performance of both stages got improved. The performance gain of the stage-one decoder is an absolute 0.8 AP score. Notably, the performance on small objects has been improved by 1.6 AP, which is a relatively large margin compared with the strong baseline. Furthermore, the improvement of the stage-one decoder also brings positive feedback to the stage-two decoder and leads to the highest performance of our SALT-Net (i.e., 42.8 AP).
\begin{figure}[htbp]
    \centering
    \begin{minipage}[1]{0.95\linewidth}
		\centerline{\includegraphics[width=\textwidth]{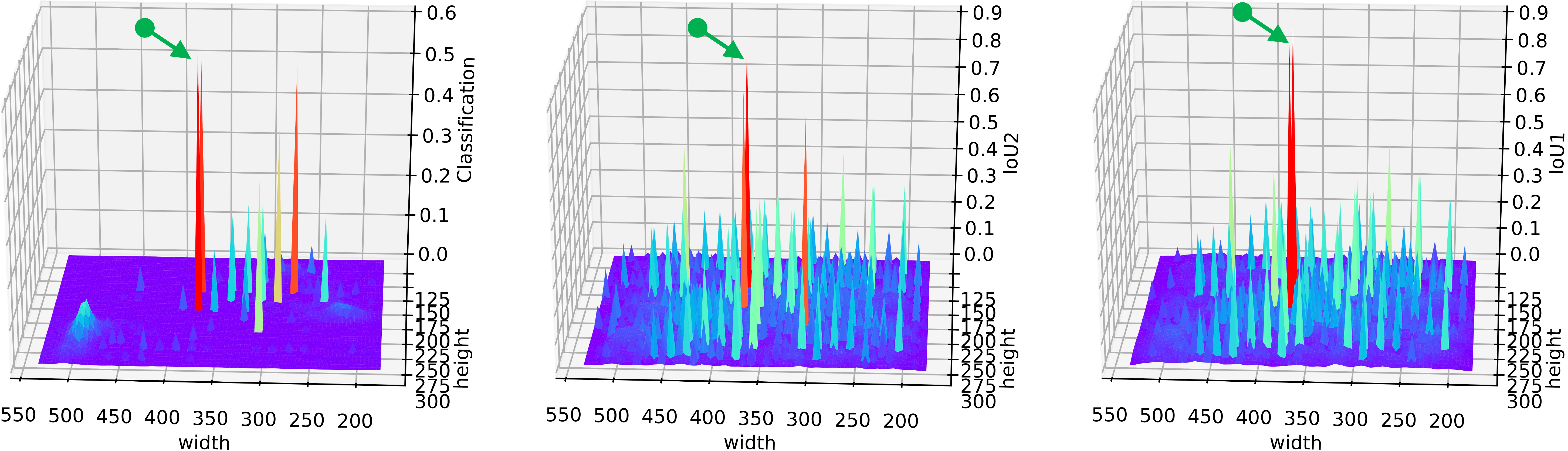}}
      \end{minipage}

      \leftline{\qquad \quad \  (a)Classification\qquad \qquad \qquad   (b)S2\qquad \qquad  \qquad \qquad\qquad (c)S1+SDR}
  
    \caption{Prediction distributions (Figure \ref{alignment0}) after applying SALT and SDR loss. The locations of their distribution peaks are identical.}
    \label{refined}
  \end{figure} 
\subsection{Evaluations for Task-alignment of SALT-Net}  
\begin{figure*}[htbp]
        \vspace{-3mm}
		\centering
		\small
		\setlength{\tabcolsep}{0.5mm}{
			\renewcommand\arraystretch{0.55}
			\begin{tabular}{ccccc}
			& \multirow{5}*{\includegraphics[height=2cm]{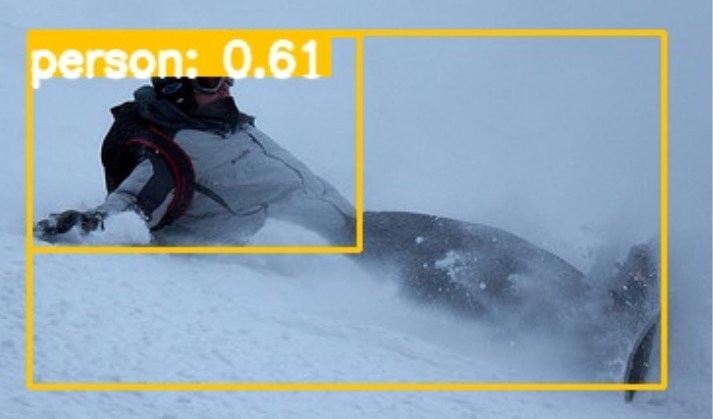}}
			& \multirow{5}*{\includegraphics[height=2cm]{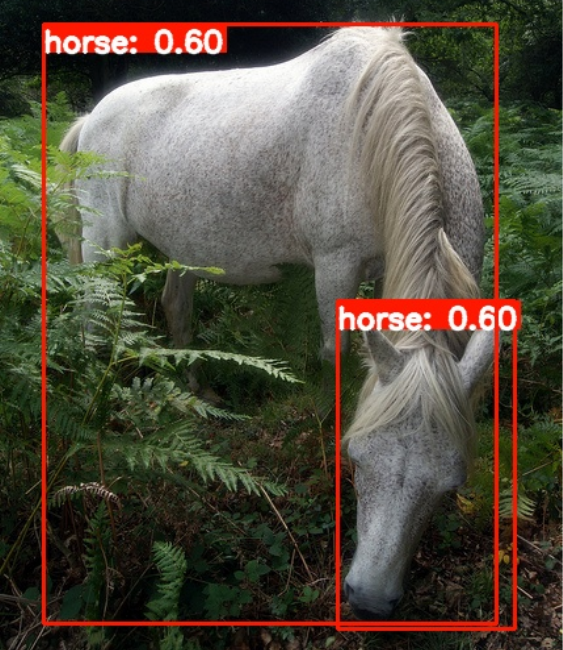}}
			& \multirow{5}*{\includegraphics[height=2cm]{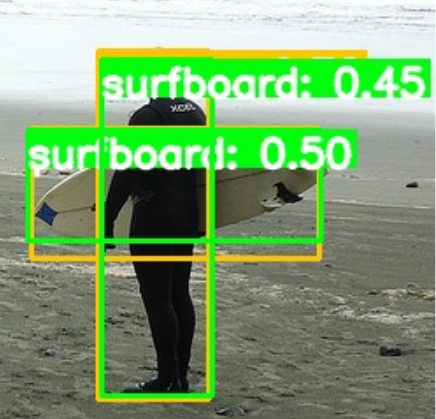}}
			& \multirow{5}*{\includegraphics[height=2cm]{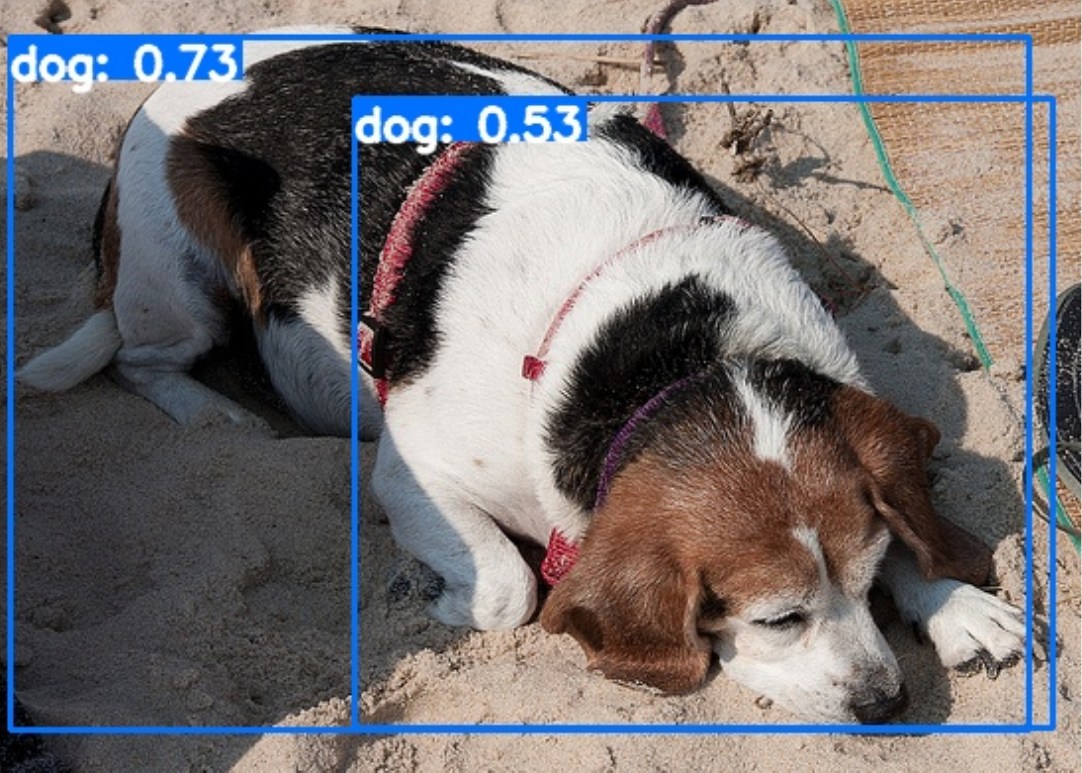}}
			~ \\~ \\~ \\ Baseline-S2 \\  \\  \\ ~ \\~ \\ ~ \\
            & \multirow{5}*{\includegraphics[height=2cm]{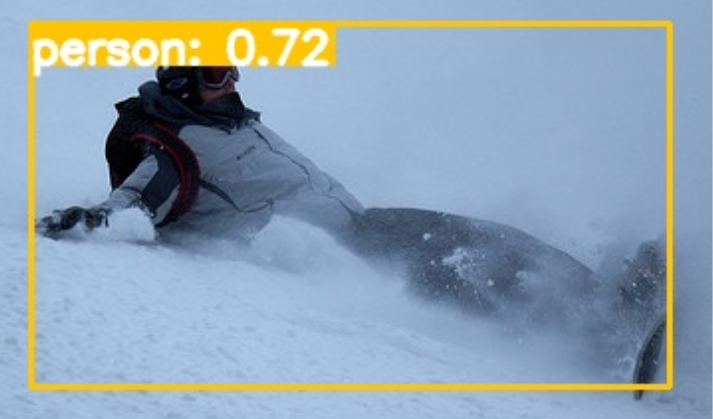}}
			& \multirow{5}*{\includegraphics[height=2cm]{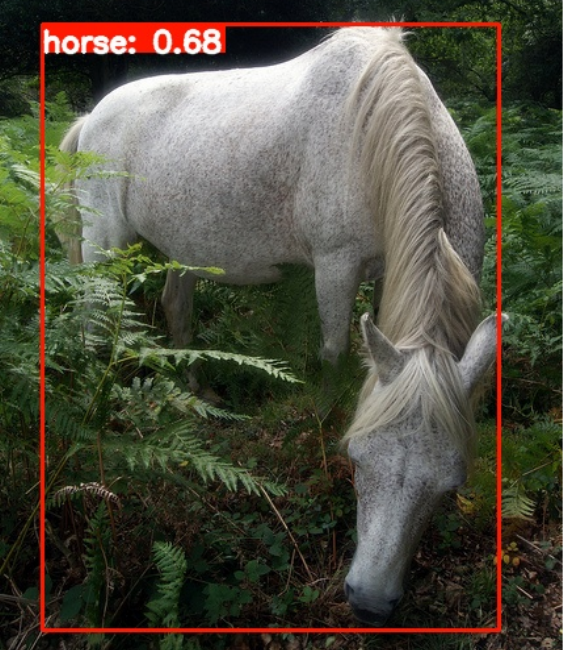}}
			& \multirow{5}*{\includegraphics[height=2cm]{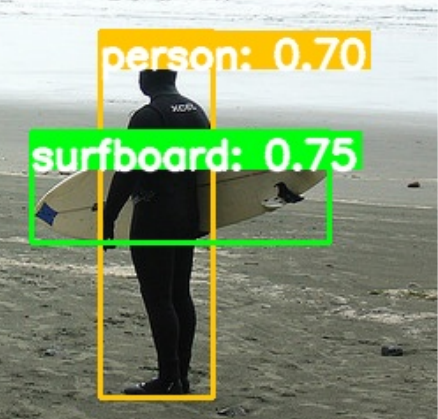}}
			& \multirow{5}*{\includegraphics[height=2cm]{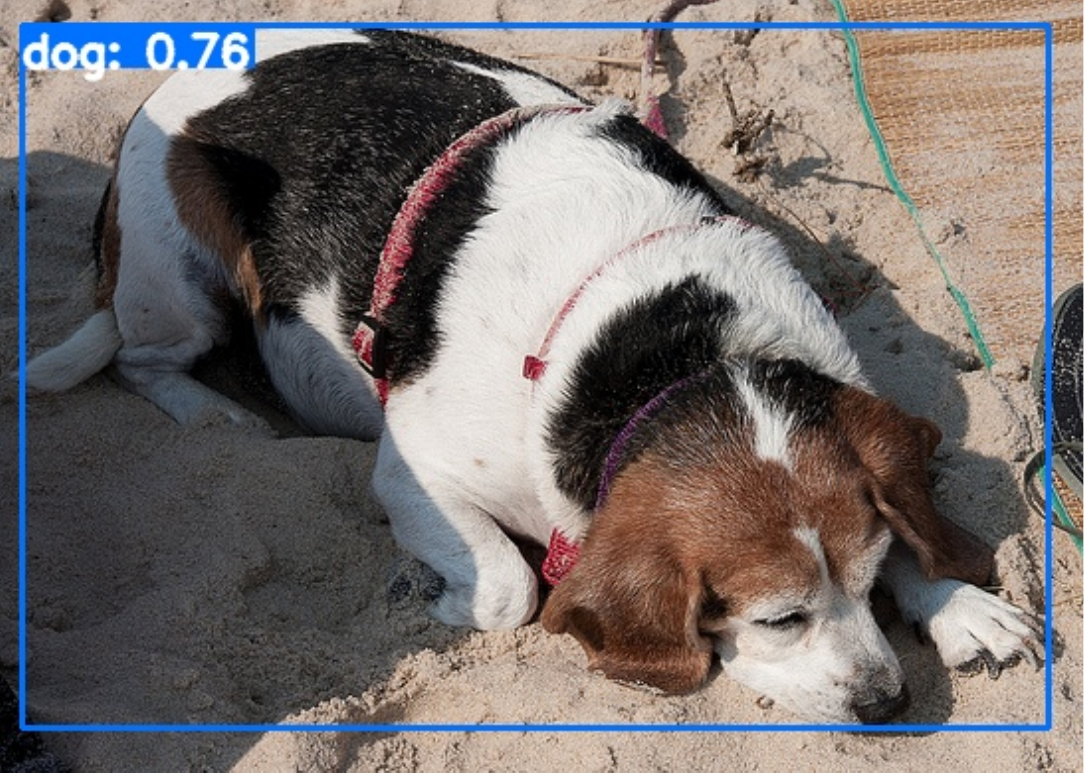}}
            ~ \\~ \\~ \\ Baseline-S2 \\ + \\SALT  \\ ~ \\~ \\ ~ \\

			\end{tabular}
		}
		\caption{Comparisons between the stage-two baseline decoder and our SALT-Net.}
		\label{qualitative}
\end{figure*}

Figure \ref{refined} (a) and (b) are the distributions of the refined detection results when implementing SALT, whereas Figure \ref{alignment0} (b) and (c) are the original coarse prediction distributions made by the stage-one decoder. The green arrows point to the distribution peaks, and one can see that they are spatially aligned (i.e., at the same location). Therefore, the detection result with the highest classification score also has the best regression result, and the misalignment gap is bridged. Figure \ref{refined} (c) is the IoU distribution of the stage-one decoder after applying the SDR loss. Its quality distributions become very close to the stage-two decoder (i.e., Figure \ref{refined} (b)), which proves the effectiveness of the regression knowledge transfer. In Figure \ref{qualitative}, the qualitative results show that SALT can align the regression and detection tasks and thereby suppress some low IoU but high classification score results.
\begin{table*}[t]
    \caption{SALT-Net {vs.} State-of-the-art Detectors. All test results are reported on the COCO {\emph{test-dev}} set.
    ``{DCN$^2$, DCN}''-applying Deformable Convolutional Network \cite{zhu2019deformable} on the last two and three stages of the backbone, respectively. ``{DCN$^{p}$}''-applying DCN on both the backbone and the FPN. ``*'' indicates applying SDR loss and ``\dag'' indicates test-time augmentations, including horizontal flip and multi-scale testing. }
    \label{sotacompare}
    \centering
    { 
    {\begin{tabular}{l|l|c|ccc|ccc} 
    \toprule
    Method&Backbone&Epoch&AP&$\mathrm{AP}_{50}$&$\mathrm{AP}_{75}$&$\mathrm{AP}_{S}$&$\mathrm{AP}_{M}$&$\mathrm{AP}_{L}$\\ 
    \hline
    \hline
    \textbf{multi-stage}\\
    GuidedAnchor\cite{wang2019region}&R-50&12&39.8&59.2&43.5&21.8&42.6&50.7\\
    DCNV2 \cite{zhu2019deformable}&X-101-32x8d-DCN&24&44.5&65.8&48.4&27.0&48.5&58.9\\
    BorderDet\cite{qiu2020borderdet}&X-101-64x4d-DCN&24&48.0&67.1&52.1&29.4&50.7&60.5\\
    RepPointsV2\cite{chen2020reppoints}&X-101-64x4d-DCN&24&49.4&68.9&53.4&30.3&52.1&62.3\\
    TSD\dag\cite{song2020revisiting}&SE154-DCN&24&51.2&71.9&56.0&33.8&54.8&64.2\\
    VFNet\cite{zhang2021varifocalnet}&X-101-32x8d-DCN&24&50.0&68.5&54.4&30.4&53.2&62.9\\
    LSNet\cite{duan2021location}&R2-101-DCN$^{p}$&24&51.1&70.3&55.2&31.2&54.3&65.1\\

    \hline
    \hline
    \textbf{one-stage}\\
    CornerNet \cite{law2018cornernet}&HG-104&200&40.5&59.1&42.3&21.8&42.7&50.2\\
    SAPD\cite{zhu2020soft}&X-101-32x8d-DCN&24&46.6&66.6&50.0&27.3&49.7&60.7\\
    ATSS\cite{zhang2020bridging}&X-101-32x8d-DCN&24&47.7&66.5&51.9&29.7&50.8&59.4\\
    GFL\cite{li2020generalized}&X-101-32x8d-DCN&24&48.2&67.4&52.6&29.2&51.7&60.2\\
    FCOS-imprv \cite{tian2020fcos}&X-101-32x8d-DCN&24&44.1&63.7&47.9&27.4&46.8&53.7\\
    PAA \cite{kim2020probabilistic}&X-101-64x4d-DCN&24&49.0&67.8&53.3&30.2&52.8&62.2\\
    GFLV2 \cite{li2021generalized}&R-50&24&44.3&62.3&48.5&26.8&47.7&54.1\\
    GFLV2 \cite{li2021generalized}&X-101-32x8d-DCN$^2$&24&49.0&67.6&53.5&29.7&52.4&61.4\\ 
    TOOD \cite{feng2021tood}&X-101-64x4d-DCN&24&51.1&69.4&55.5&31.9&54.1&63.7\\
    \hline  
    \rowcolor{blue!5} SALT-Net${^*}$&R-50&24&46.1&64.0&50.3&28.0&49.5&57.2\\
    \rowcolor{blue!5} SALT-Net&X-101-32x8d-DCN$^2$&24&49.8&68.5&54.2&30.6&53.2&62.6\\
    \rowcolor{blue!5} SALT-Net&X-101-32x8d-DCN&24&50.2&68.8&54.9&31.2&53.4&63.1\\
    \rowcolor{blue!5} SALT-Net&R2-101-DCN$^2$&24&51.1&69.7&55.7&32.3&54.5&64.0\\
    \rowcolor{blue!5} SALT-Net${^*}$&R2-101-DCN&24&51.5&70.0&56.2&32.1&55.1&64.8\\
    \rowcolor{blue!5} SALT-Net${^*_\dag}$&R2-101-DCN&24&\textbf{53.8}&\textbf{71.1}&\textbf{59.9}&\textbf{36.3}&\textbf{56.9}&\textbf{65.1}\\      
    \bottomrule
    
    \end{tabular}}} 
    
    \end{table*}

  \subsection{Comparisons with State-of-the-arts}
  The multi-scale training strategy (i.e., input images are resized from [400, 1333] to [960, 1333]) and the 2 $\times$ schedule \cite{chen2019mmdetection} are adopted as they are commonly used strategies in state-of-the-art methods. GFLV2 only applies DCN on the last two stages of the backbone, whereas the common practices \cite{tian2020fcos,zhang2021varifocalnet} usually apply it on the last three stages. Therefore, for a fair comparison, the results of the proposed method with both settings are reported. As Table \ref{sotacompare} shows, our model achieves a 46.1 AP with ResNet-50, which outperforms other state-of-the-art methods with heavier backbones (e.g., FCOS with X-101-32x8d-DCN). With test-time augmentations and R2-101-DCN as the backbone, our best model achieves a 53.8 AP, which is a very competitive result among dense object detectors.
   
  \section{Conclusion} 
  In this work, we presented SALT, a simple yet effective plug-in operator that can solve the misalignment problem between regression and classification. Our new-fashioned framework can disentangle classification and regression from the spatial dimension by extracting features from each task's sensitive locations and aligning them to the same anchor point. We also proposed SDR loss to transfer the regression knowledge from the stage-two decoder to the stage-one decoder. The refined detection results also received positive feedback by improving the coarse regression results, and the final performance improved in an inference cost-free fashion. Extensive experiments showed that SALT could considerably raise the performance of various dense detectors, and SALT-Net showed promising results among the state-of-the-art dense detectors.
\bibliographystyle{splncs04}
\bibliography{mybibliography}

\begin{thebibliography}{10}\itemsep=-1pt

\bibitem{azulay2019deep}
Aharon Azulay and Yair Weiss.
\newblock Why do deep convolutional networks generalize so poorly to small
  image transformations?
\newblock {\em Journal of Machine Learning Research}, 20:1--25, 2019.

\bibitem{chen2020reppoints}
Yihong Chen, Zheng Zhang, Yue Cao, Liwei Wang, Stephen Lin, and Han Hu.
\newblock Reppoints v2: Verification meets regression for object detection.
\newblock {\em Advances in Neural Information Processing Systems}, 33, 2020.

\bibitem{islam2020much}
Md~Amirul Islam, Sen Jia, and Neil~DB Bruce.
\newblock How much position information do convolutional neural networks
  encode?
\newblock {\em arXiv preprint arXiv:2001.08248}, 2020.

\bibitem{kim2020probabilistic}
Kang Kim and Hee~Seok Lee.
\newblock Probabilistic anchor assignment with iou prediction for object
  detection.
\newblock In {\em ECCV}, pages 355--371. Springer, 2020.

\bibitem{li2021generalized}
Xiang Li, Wenhai Wang, Xiaolin Hu, Jun Li, Jinhui Tang, and Jian Yang.
\newblock Generalized focal loss v2: Learning reliable localization quality
  estimation for dense object detection.
\newblock In {\em CVPR}, pages 11632--11641, 2021.

\bibitem{li2020generalized}
Xiang Li, Wenhai Wang, Lijun Wu, Shuo Chen, Xiaolin Hu, Jun Li, Jinhui Tang,
  and Jian Yang.
\newblock Generalized focal loss: Learning qualified and distributed bounding
  boxes for dense object detection.
\newblock In {\em NeurIPS}, 2020.

\bibitem{lin2014microsoft}
Tsung-Yi Lin, Michael Maire, Serge Belongie, James Hays, Pietro Perona, Deva
  Ramanan, Piotr Doll{\'a}r, and C~Lawrence Zitnick.
\newblock Microsoft coco: Common objects in context.
\newblock In {\em ECCV}, pages 740--755. Springer, 2014.

\bibitem{qiu2020borderdet}
Han Qiu, Yuchen Ma, Zeming Li, Songtao Liu, and Jian Sun.
\newblock Borderdet: Border feature for dense object detection.
\newblock In {\em ECCV}, pages 549--564. Springer, 2020.

\bibitem{tian2019fcos}
Zhi Tian, Chunhua Shen, Hao Chen, and Tong He.
\newblock Fcos: Fully convolutional one-stage object detection.
\newblock In {\em ICCV}, pages 9627--9636, 2019.

\bibitem{zhang2020bridging}
Shifeng Zhang, Cheng Chi, Yongqiang Yao, Zhen Lei, and Stan~Z Li.
\newblock Bridging the gap between anchor-based and anchor-free detection via
  adaptive training sample selection.
\newblock In {\em CVPR}, pages 9759--9768, 2020.

\bibitem{zhu2020soft}
Chenchen Zhu, Fangyi Chen, Zhiqiang Shen, and Marios Savvides.
\newblock Soft anchor-point object detection.
\newblock In {\em ECCV}, pages 91--107. Springer, 2020.

\end{thebibliography}


\begin{thebibliography}{10}
\providecommand{\url}[1]{\texttt{#1}}
\providecommand{\urlprefix}{URL }
\providecommand{\doi}[1]{https://doi.org/#1}

\bibitem{azulay2019deep}
Azulay, A., Weiss, Y.: Why do deep convolutional networks generalize so poorly
  to small image transformations? Journal of Machine Learning Research
  \textbf{20},  1--25 (2019)

\bibitem{chen2019mmdetection}
Chen, K., Wang, J., Pang, J., Cao, Y., Xiong, Y., Li, X., Sun, S., Feng, W.,
  Liu, Z., Xu, J., et~al.: Mmdetection: Open mmlab detection toolbox and
  benchmark. arXiv preprint arXiv:1906.07155  (2019)

\bibitem{chen2020reppoints}
Chen, Y., Zhang, Z., Cao, Y., Wang, L., Lin, S., Hu, H.: Reppoints v2:
  Verification meets regression for object detection. Advances in Neural
  Information Processing Systems  \textbf{33} (2020)

\bibitem{chi2019selective}
Chi, C., Zhang, S., Xing, J., Lei, Z., Li, S.Z., Zou, X.: Selective refinement
  network for high performance face detection. In: AAAI. vol.~33, pp.
  8231--8238 (2019)

\bibitem{dai2017deformable}
Dai, J., Qi, H., Xiong, Y., Li, Y., Zhang, G., Hu, H., Wei, Y.: Deformable
  convolutional networks. In: ICCV. pp. 764--773 (2017)

\bibitem{duan2021location}
Duan, K., Xie, L., Qi, H., Bai, S., Huang, Q., Tian, Q.: Location-sensitive
  visual recognition with cross-iou loss. arXiv preprint arXiv:2104.04899
  (2021)

\bibitem{feng2021tood}
Feng, C., Zhong, Y., Gao, Y., Scott, M.R., Huang, W.: Tood: Task-aligned
  one-stage object detection. In: 2021 IEEE/CVF International Conference on
  Computer Vision (ICCV). pp. 3490--3499. IEEE Computer Society (2021)

\bibitem{hinton2015distilling}
Hinton, G., Vinyals, O., Dean, J., et~al.: Distilling the knowledge in a neural
  network. arXiv preprint arXiv:1503.02531  \textbf{2}(7) (2015)

\bibitem{islam2020much}
Islam, M.A., Jia, S., Bruce, N.D.: How much position information do
  convolutional neural networks encode? arXiv preprint arXiv:2001.08248  (2020)

\bibitem{ji2021refine}
Ji, M., Shin, S., Hwang, S., Park, G., Moon, I.C.: Refine myself by teaching
  myself: Feature refinement via self-knowledge distillation. In: CVPR. pp.
  10664--10673 (2021)

\bibitem{kang2021instance}
Kang, Z., Zhang, P., Zhang, X., Sun, J., Zheng, N.: Instance-conditional
  knowledge distillation for object detection. NeurIPS  \textbf{34} (2021)

\bibitem{kim2020probabilistic}
Kim, K., Lee, H.S.: Probabilistic anchor assignment with iou prediction for
  object detection. In: ECCV. pp. 355--371. Springer (2020)

\bibitem{law2018cornernet}
Law, H., Deng, J.: Cornernet: Detecting objects as paired keypoints. In: ECCV.
  pp. 734--750 (2018)

\bibitem{li2021generalized}
Li, X., Wang, W., Hu, X., Li, J., Tang, J., Yang, J.: Generalized focal loss
  v2: Learning reliable localization quality estimation for dense object
  detection. In: CVPR. pp. 11632--11641 (2021)

\bibitem{li2020generalized}
Li, X., Wang, W., Wu, L., Chen, S., Hu, X., Li, J., Tang, J., Yang, J.:
  Generalized focal loss: Learning qualified and distributed bounding boxes for
  dense object detection. In: NeurIPS (2020)

\bibitem{li2021student}
Li, Z., Li, X., Yang, L., Yang, J., Pan, Z.: Student helping teacher: Teacher
  evolution via self-knowledge distillation. arXiv preprint arXiv:2110.00329
  (2021)

\bibitem{lin2017feature}
Lin, T.Y., Doll{\'a}r, P., Girshick, R., He, K., Hariharan, B., Belongie, S.:
  Feature pyramid networks for object detection. In: CVPR. pp. 2117--2125
  (2017)

\bibitem{lin2017focal}
Lin, T.Y., Goyal, P., Girshick, R., He, K., Doll{\'a}r, P.: Focal loss for
  dense object detection. In: ICCV. pp. 2980--2988 (2017)

\bibitem{lin2014microsoft}
Lin, T.Y., Maire, M., Belongie, S., Hays, J., Perona, P., Ramanan, D.,
  Doll{\'a}r, P., Zitnick, C.L.: Microsoft coco: Common objects in context. In:
  ECCV. pp. 740--755. Springer (2014)

\bibitem{qiu2020borderdet}
Qiu, H., Ma, Y., Li, Z., Liu, S., Sun, J.: Borderdet: Border feature for dense
  object detection. In: ECCV. pp. 549--564. Springer (2020)

\bibitem{redmon2016you}
Redmon, J., Divvala, S., Girshick, R., Farhadi, A.: You only look once:
  Unified, real-time object detection. In: CVPR. pp. 779--788 (2016)

\bibitem{yolov3}
Redmon, J., Farhadi, A.: Yolov3: An incremental improvement. arXiv  (2018)

\bibitem{rezatofighi2019generalized}
Rezatofighi, H., Tsoi, N., Gwak, J., Sadeghian, A., Reid, I., Savarese, S.:
  Generalized intersection over union: A metric and a loss for bounding box
  regression. In: CVPR. pp. 658--666 (2019)

\bibitem{song2020revisiting}
Song, G., Liu, Y., Wang, X.: Revisiting the sibling head in object detector.
  In: CVPR. pp. 11563--11572 (2020)

\bibitem{tian2019fcos}
Tian, Z., Shen, C., Chen, H., He, T.: Fcos: Fully convolutional one-stage
  object detection. In: ICCV. pp. 9627--9636 (2019)

\bibitem{tian2020fcos}
Tian, Z., Shen, C., Chen, H., He, T.: Fcos: A simple and strong anchor-free
  object detector. IEEE Transactions on Pattern Analysis and Machine
  Intelligence  (2020)

\bibitem{wang2019region}
Wang, J., Chen, K., Yang, S., Loy, C.C., Lin, D.: Region proposal by guided
  anchoring. In: CVPR. pp. 2965--2974 (2019)

\bibitem{wang2019distilling}
Wang, T., Yuan, L., Zhang, X., Feng, J.: Distilling object detectors with
  fine-grained feature imitation. In: CVPR. pp. 4933--4942 (2019)

\bibitem{wu2020iou}
Wu, S., Li, X., Wang, X.: Iou-aware single-stage object detector for accurate
  localization. Image and Vision Computing  \textbf{97},  103911 (2020)

\bibitem{wu2020rethinking}
Wu, Y., Chen, Y., Yuan, L., Liu, Z., Wang, L., Li, H., Fu, Y.: Rethinking
  classification and localization for object detection. In: CVPR. pp.
  10186--10195 (2020)

\bibitem{yang2019reppoints}
Yang, Z., Liu, S., Hu, H., Wang, L., Lin, S.: Reppoints: Point set
  representation for object detection. In: ICCV. pp. 9657--9666 (2019)

\bibitem{yao2020knowledge}
Yao, A., Sun, D.: Knowledge transfer via dense cross-layer mutual-distillation.
  In: ECCV. pp. 294--311. Springer (2020)

\bibitem{zhang2021varifocalnet}
Zhang, H., Wang, Y., Dayoub, F., Sunderhauf, N.: Varifocalnet: An iou-aware
  dense object detector. In: CVPR. pp. 8514--8523 (2021)

\bibitem{zhang2019your}
Zhang, L., Song, J., Gao, A., Chen, J., Bao, C., Ma, K.: Be your own teacher:
  Improve the performance of convolutional neural networks via self
  distillation. In: ICCV. pp. 3713--3722 (2019)

\bibitem{zhang2021lgd}
Zhang, P., Kang, Z., Yang, T., Zhang, X., Zheng, N., Sun, J.: Lgd: Label-guided
  self-distillation for object detection. arXiv preprint arXiv:2109.11496
  (2021)

\bibitem{zhang2020bridging}
Zhang, S., Chi, C., Yao, Y., Lei, Z., Li, S.Z.: Bridging the gap between
  anchor-based and anchor-free detection via adaptive training sample
  selection. In: CVPR. pp. 9759--9768 (2020)

\bibitem{zhang2018single}
Zhang, S., Wen, L., Bian, X., Lei, Z., Li, S.Z.: Single-shot refinement neural
  network for object detection. In: CVPR. pp. 4203--4212 (2018)

\bibitem{zhang2018deep}
Zhang, Y., Xiang, T., Hospedales, T.M., Lu, H.: Deep mutual learning. In: CVPR.
  pp. 4320--4328 (2018)

\bibitem{zhu2020soft}
Zhu, C., Chen, F., Shen, Z., Savvides, M.: Soft anchor-point object detection.
  In: ECCV. pp. 91--107. Springer (2020)

\bibitem{zhu2019deformable}
Zhu, X., Hu, H., Lin, S., Dai, J.: Deformable convnets v2: More deformable,
  better results. In: CVPR. pp. 9308--9316 (2019)

\end{thebibliography}
\clearpage
\appendix

\section{Introduction}
In this supplementary material, we provide the inference details of SALT-Net and the skip connection experiment on the regression and classification towers. Finally, we demonstrate the images of the detection results of SALT-Net on the MS-COCO \cite{lin2014microsoft} val2017 split.
\begin{figure*}[t]
  \centering
  \includegraphics[width=0.99\textwidth]{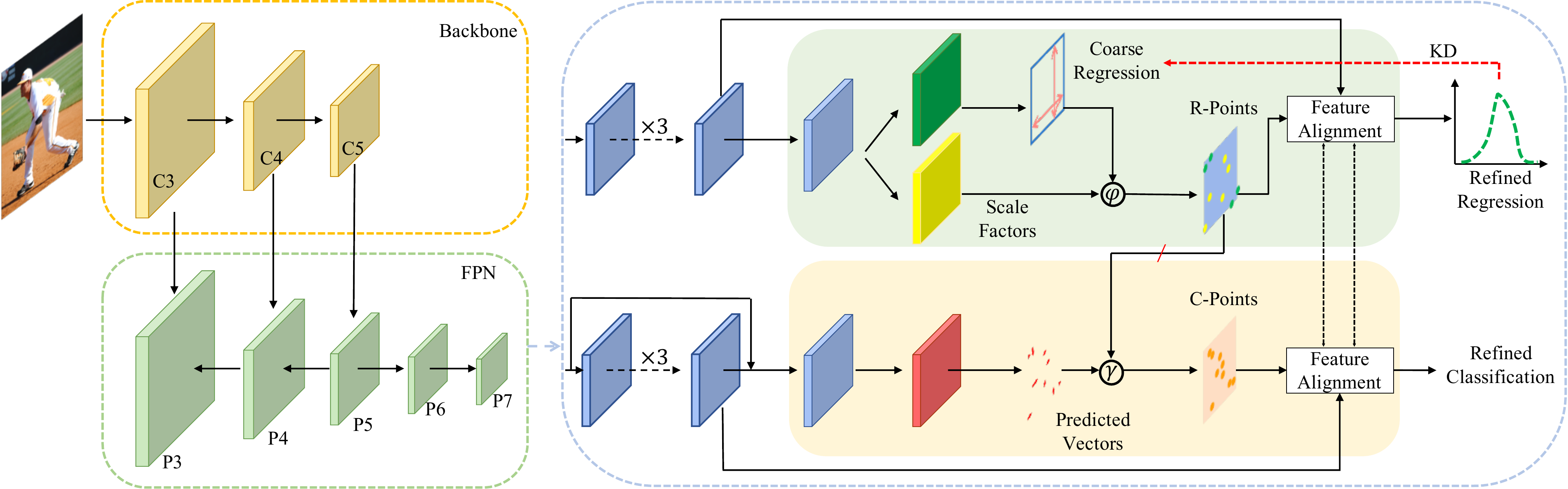} 
  \caption{Architecture of SALT-Net. }
  \label{networka} 
  \end{figure*}

  \section{Network Architecture and Inference}

  Figure \ref{networka} presents the network of our proposed SALT-Net. The input features from FPN are fed into two parallel subnetworks for the regression and classification tasks. We also apply a skip connection on the classification tower because we find that it can improve the accuracy, and the ablation study is in Sec. \ref{spc}. 

  First, the regression subnetwork predicts the coarse regression result with misaligned features. Based on the coarse bounding box, SALT predicts the normalized scale factors $\mathcal S$ and then outputs a set of regression-aware points with Equation (Sec 3.1: 1,2). Features from the locations of the regression-aware points are extracted for predicting the refined regression result.
  
  By taking the regression-aware points as the point-set anchor, SALT predicts the disentanglement vectors $\mathcal D$ and then outputs the classification-aware points with Equation (Sec 3.2: 5). After extracting the aligned classification features from the locations of the classification-aware points, the classification subnetwork outputs the refined classification result. 
  
 Finally, the detection results from all levels of FPN are merged, and we use NMS to filter out redundant results with the threshold set as 0.6.
 
\subsection{Inference Speed}
\begin{figure}[t]
  \centering
  \includegraphics[width=0.6\linewidth]{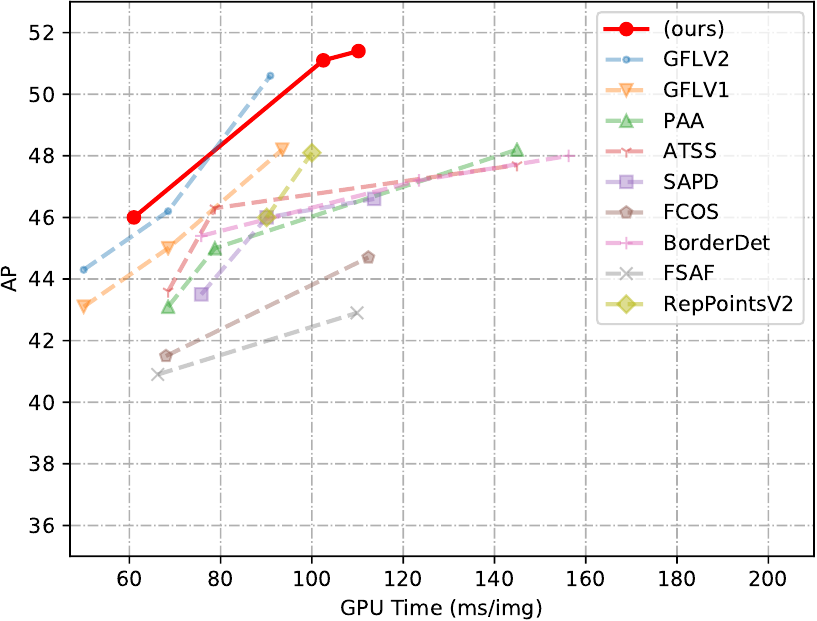}
  \caption{Single-model single-scale speed (ms) vs. accuracy (AP) on COCO \emph{test-dev} among state-of-the-art dense object detectors.}
  \label{speed} 
  \end{figure} 
SALT-Net is a dense object detector that predicts the detection result in a per-pixel prediction fashion. Figure \ref{speed} illustrates the comparisons between the SALT-Net and other state-of-the-art dense object detectors \cite{li2020generalized,li2021generalized,kim2020probabilistic,zhang2020bridging,zhu2020soft,tian2019fcos,qiu2020borderdet,chen2020reppoints}. All test results are reported with a single Titan RTX GPU and a Xeon Gold 6230 CPU. Notably, our prosed approach achieves a new level of accuracy-speed trade-off.

\section{Skip Connection}\label{spc}
We did an experiment about applying a long-range skip connection (i.e., LS) on the classification and regression subnetworks. Table \ref{tab:skip} is the experiment result, and it revealed an interesting finding that the effect of LS on the two tasks is quite the opposite. Here are our explanations for this interesting finding and our motivation for doing this experiment.

Classification has translation and scale invariance, whereas regression is the contrary. The “stride”  and “zero-padding” of the convolution operation can also affect the invariance property. However, the cumulative number of stride and padding times varied in feature maps with different depths of the CNNs. Therefore, we want to explore how the feature fusion would affect the two tasks, respectively. The LS is used in our investigation to fuse feature maps with different depths of the network. As shown in Table \ref{tab:skip}, the implementation of the LS on the classification tower can improve the AP by 0.4. Nevertheless, applying LS on the regression tower decreases the AP by 0.3. That indicates that the effect of the LS on these two tasks is the opposite. 
   \begin{table}[t]
    \caption{Different skip connection strategies. ``C'' and ``R'' denote classification and regression tower, respectively. We have tried three kinds of strategies: applying the long-range skip connection on the regression and the classification tower, respectively, and without the skip connection in both subnetworks.}
      \centering
      \footnotesize
      {\begin{tabular}{cc|c|cc|ccc} 
      \toprule
      C&R&$\mathrm{AP}$&$\mathrm{AP}_{50}$&$\mathrm{AP}_{75}$&$\mathrm{AP}_{S}$&$\mathrm{AP}_{M}$&$\mathrm{AP}_{L}$\\ 
      \hline
      \hline
      &&42.1&59.6&45.6&24.8&45.4&55.5 \\
      &\checkmark&41.8&59.4&45.6&23.6&45.6&55.6   \\
      \checkmark&&\textbf{42.5}&\textbf{60.1}&\textbf{46.2}&\textbf{25.1}&\textbf{45.9}&\textbf{56.4}     \\
      \bottomrule
      \end{tabular}}
     
      \label{tab:skip} 
      \end{table}
\subsection{Motivation for the Skip Connection Experiment}
  \begin{enumerate}
    \item Classification has translation and scale invariance, while regression is quite the opposite.
\item Feature maps that come from different depths of the network have different impacts on the invariance property \cite{azulay2019deep}.
\item There is a direct correlation between the regression task and the depth of the feature map \cite{islam2020much}.
\item The skip connection is almost a cost-free method that can fuse feature maps with different depths.
\end{enumerate}
\subsection{Details}
\begin{figure}[htbp]
  \centering
  \begin{subfigure}{0.32\linewidth}\centering
    \includegraphics[width=0.6\textwidth]{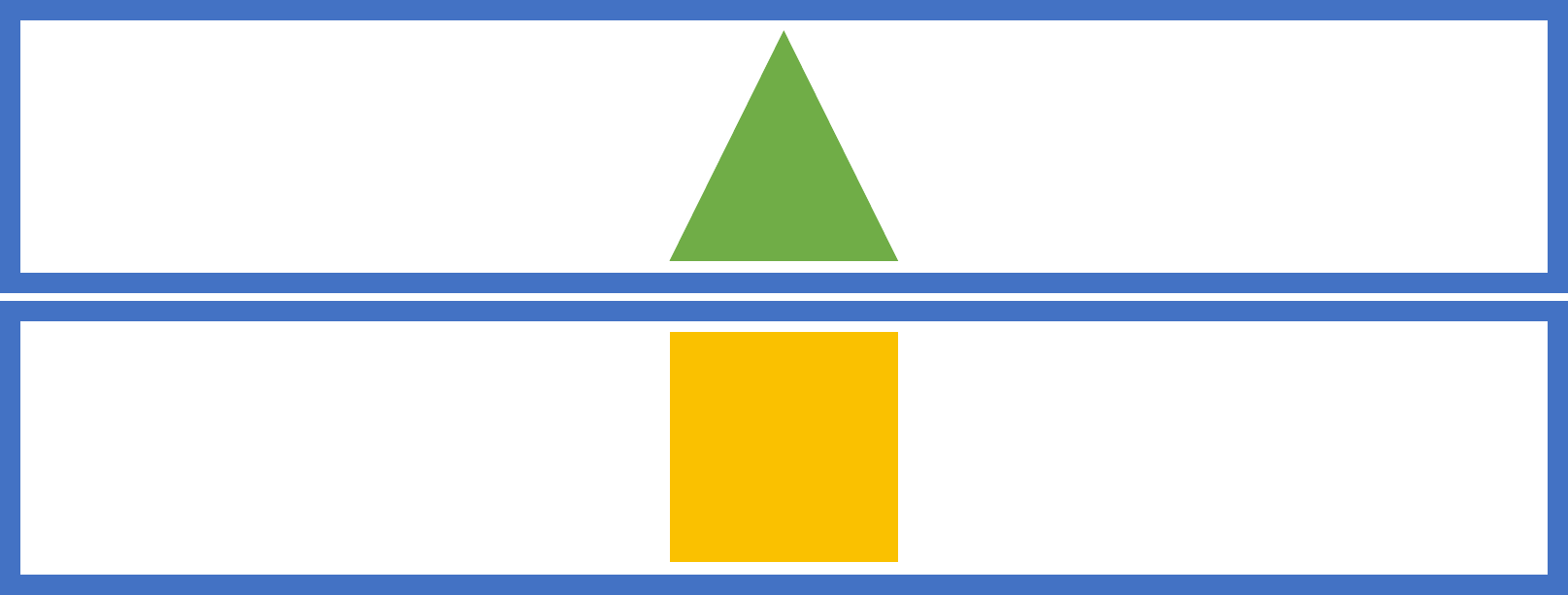}
    \caption{}
  \end{subfigure}
\begin{subfigure}{0.32\linewidth}\centering
    \includegraphics[width=0.6\textwidth]{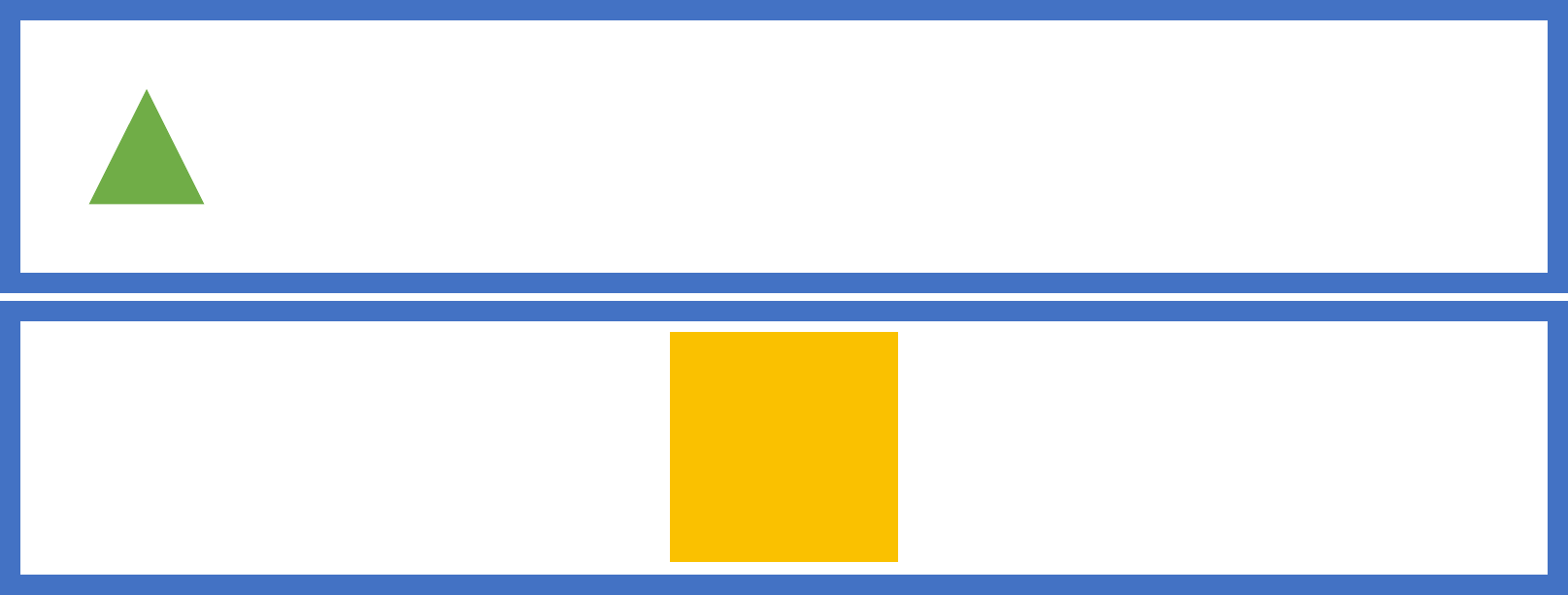}
    \caption{invariance}
  \end{subfigure}
  \begin{subfigure}{0.32\linewidth}\centering
    \includegraphics[width=0.6\textwidth]{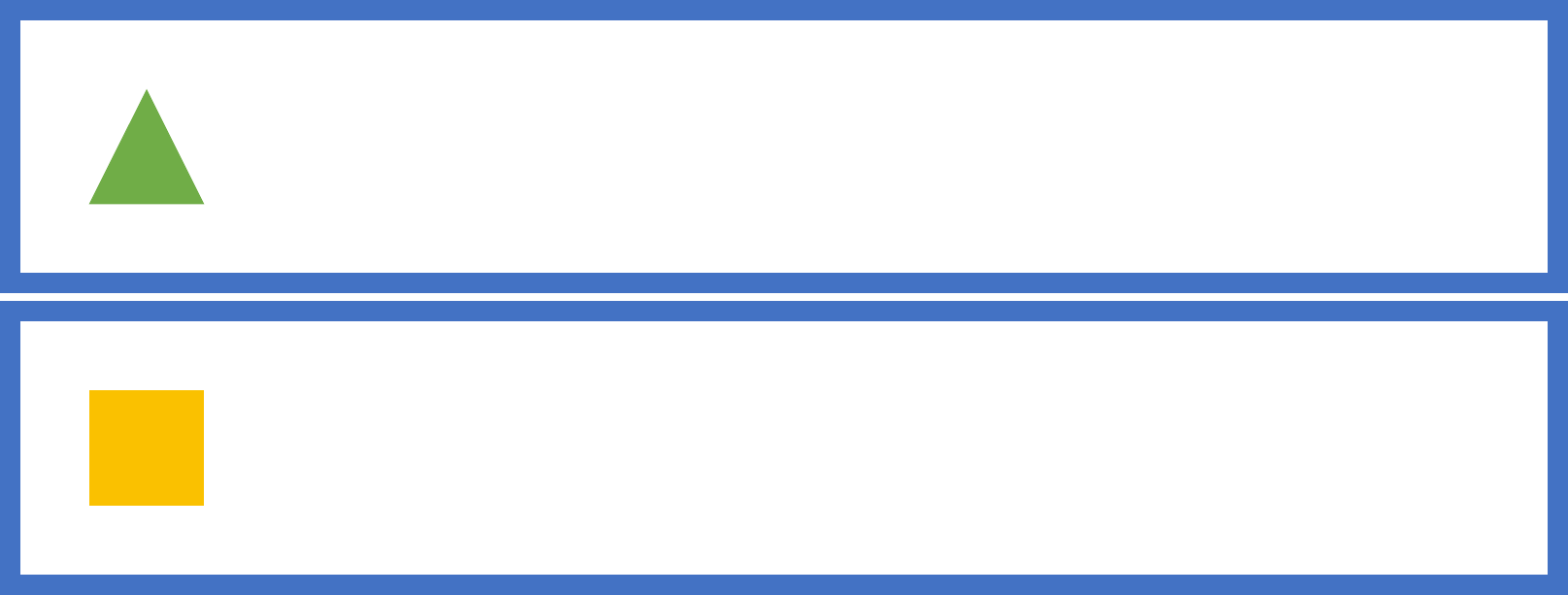}
    \caption{equivariance}
  \end{subfigure}

   \caption{(a) is the control group. The input images of (b) and (c) have the location and size transformations.}
  \label{invariance} 
  \end{figure}

As illustrated in Figure \ref{invariance}, the triangles and rectangles represent the input images and their output features, respectively. 
Classification has translation and scale invariance, that is, the location and size transformations of the object do not affect the classification result (i.e., Figure \ref{invariance} (b)). Nevertheless, regression has translation and scale equivariance, that is, the location and size transformations of the object have the same effect on the regression result (i.e., Figure \ref{invariance} (c)).

  \begin{figure}[htbp]
    \centering
    \includegraphics[width=0.6\linewidth]{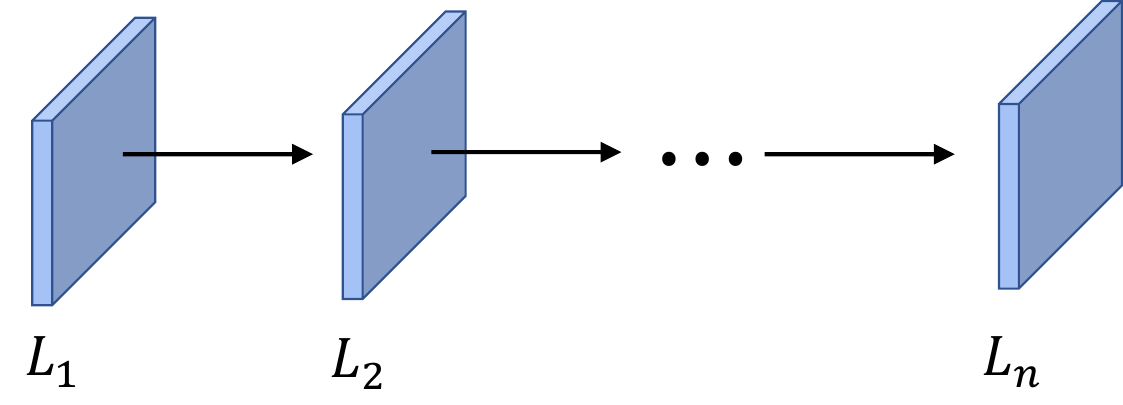} 
        \caption{Feature maps with different "depths". }
    \label{ln} 
    \end{figure} 

Our motivation for doing the skip-connection experiment comes from two studies. First, Convolutional Neural Networks (CNNs) are assumed to be invariant to scale and translation changes. However, Azulay and Weiss \cite{azulay2019deep} argued that it is not the case. The \textcolor[rgb]{1,0,0}{subsampling} and \textcolor[rgb]{1,0,0}{stride} of the convolution operation will affect the invariance. The deeper the network, the more significant it is, especially for small objects. Let $f$ be the function that measures the cumulative number of subsampling and stride times of feature map $L_n$. $f$ is proportional to the depth of the feature map $n$, as Equation \ref{fn} shows.
\begin{equation}
  f(L_n)\propto n
  \label{fn}
  \end{equation}

  Second, Islam et al. \cite{islam2020much} revealed that the reason why CNNs can learn absolute position is because of the commonly used \textcolor[rgb]{1,0,0}{zero-padding} operation. However, the padding times of feature maps with different depths are different. Let $g$ be the function that measures the cumulative number of zero-padding times of feature map $L_n$. $g$ is also proportional to the depth of the feature map $n$, as Equation \ref{gn} shows.
\begin{equation}
  g(L_n)\propto n
  \label{gn}
  \end{equation}

\begin{figure*}[htbp]
  \centering
  \begin{subfigure}{0.32\linewidth}
    \includegraphics[width=0.99\textwidth]{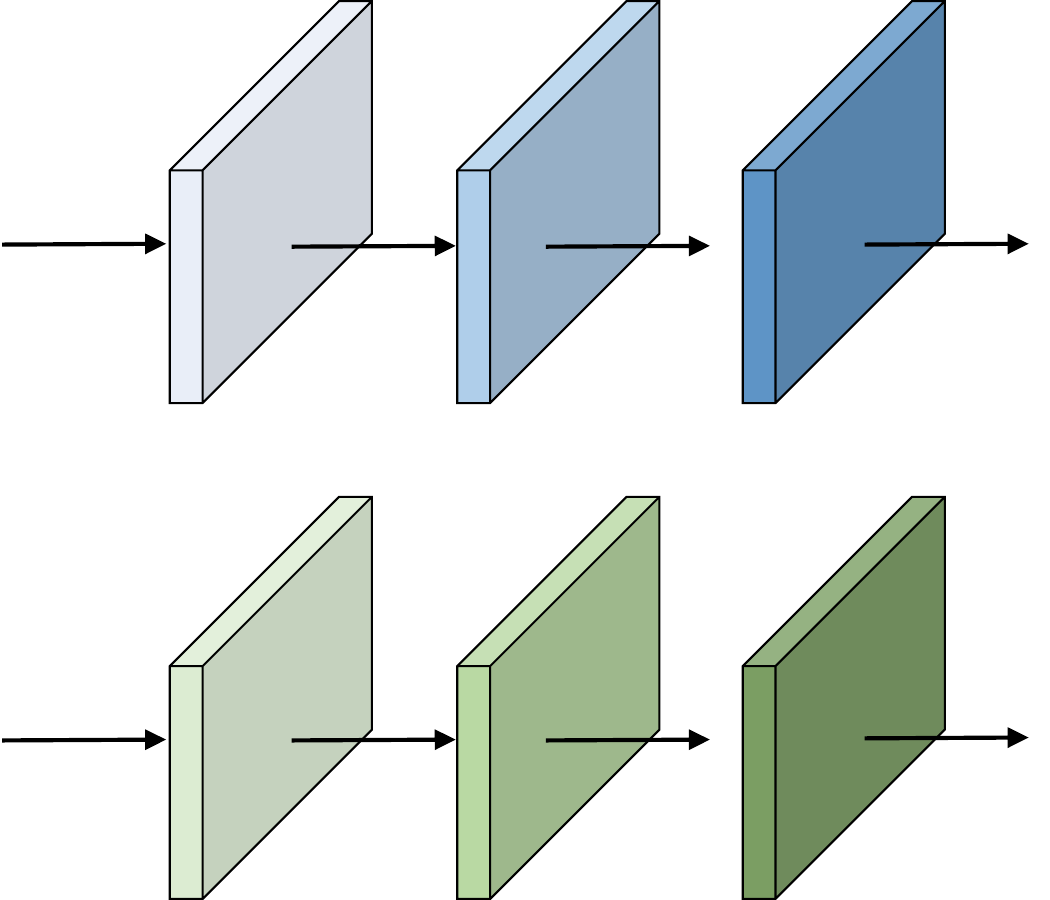}
    \caption{without LS}
  \end{subfigure}
\begin{subfigure}{0.32\linewidth}
    \includegraphics[width=0.99\textwidth]{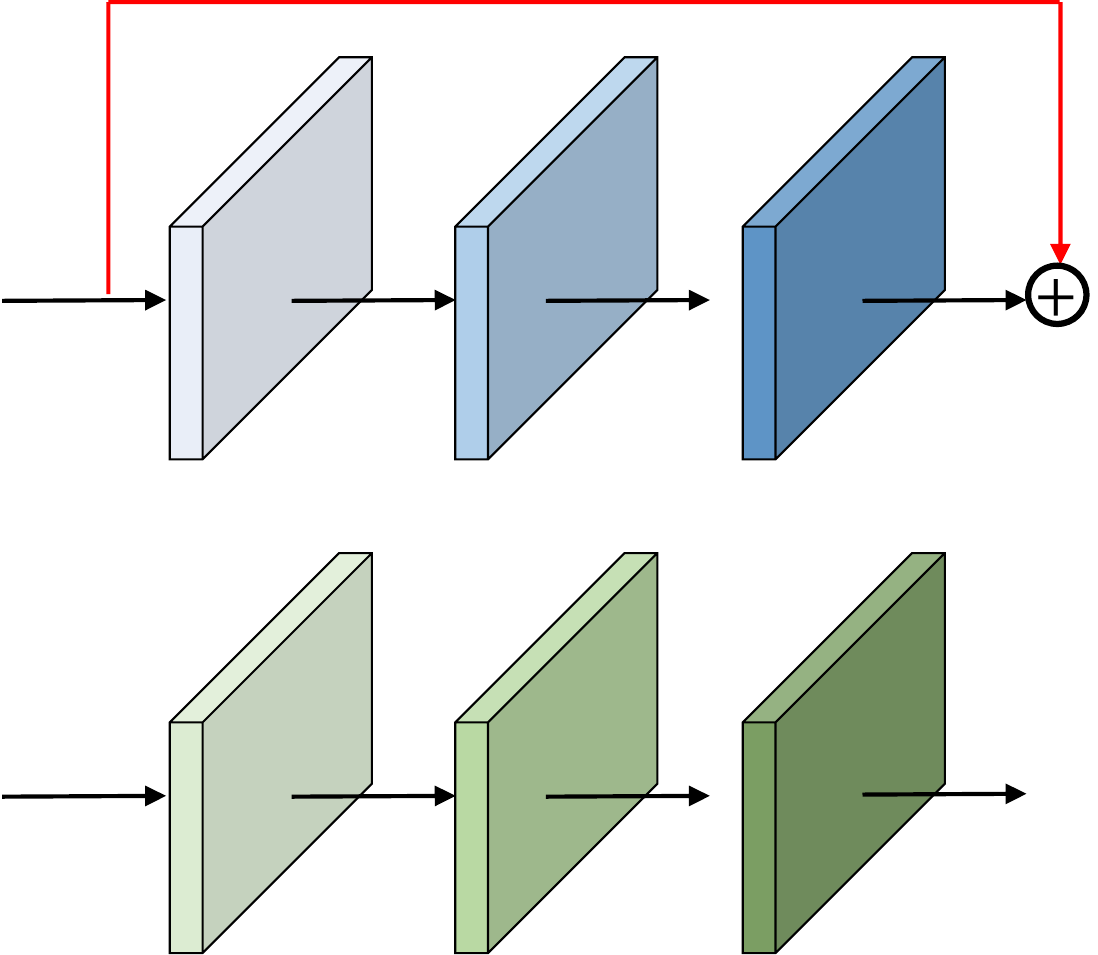}
    \caption{LS-regression}
  \end{subfigure}
  \begin{subfigure}{0.32\linewidth}
    \includegraphics[width=0.99\textwidth]{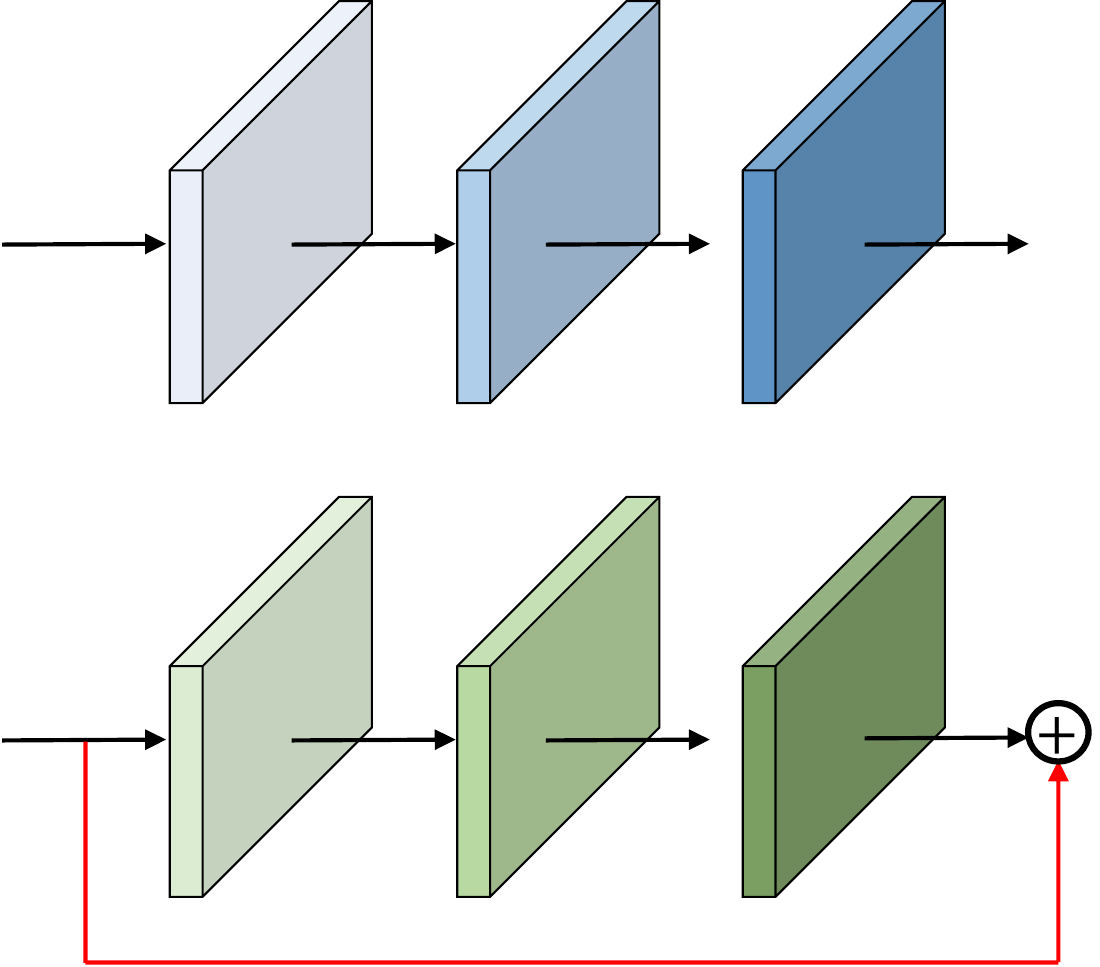}
    \caption{LS-classification}
  \end{subfigure}

 \caption{Illustration of different long-range skip connection strategies. (b) and (c) denote the regression and classification subnetworks, respectively. }
  \label{skip} 
  \end{figure*}

  The skip connection can affect the depth of the feature map. Therefore, it also has an influence on the classification and the position decoding task (regression). As illustrated in Figure \ref{skip}, we use LS to fuse features maps from different depths of the network. The deeper the color, the deeper the feature map is. Experiments in table \ref{tab:skip} show that the effect of the LS for the two tasks is quite the opposite. The feature fusion process is beneficial to the classification task while harmful to the regression result.

  \section{Refined Regression Results}
\begin{figure*}[htbp]
    \centering
    \vspace{-7.5cm}
    \includegraphics[width=0.99\textwidth]{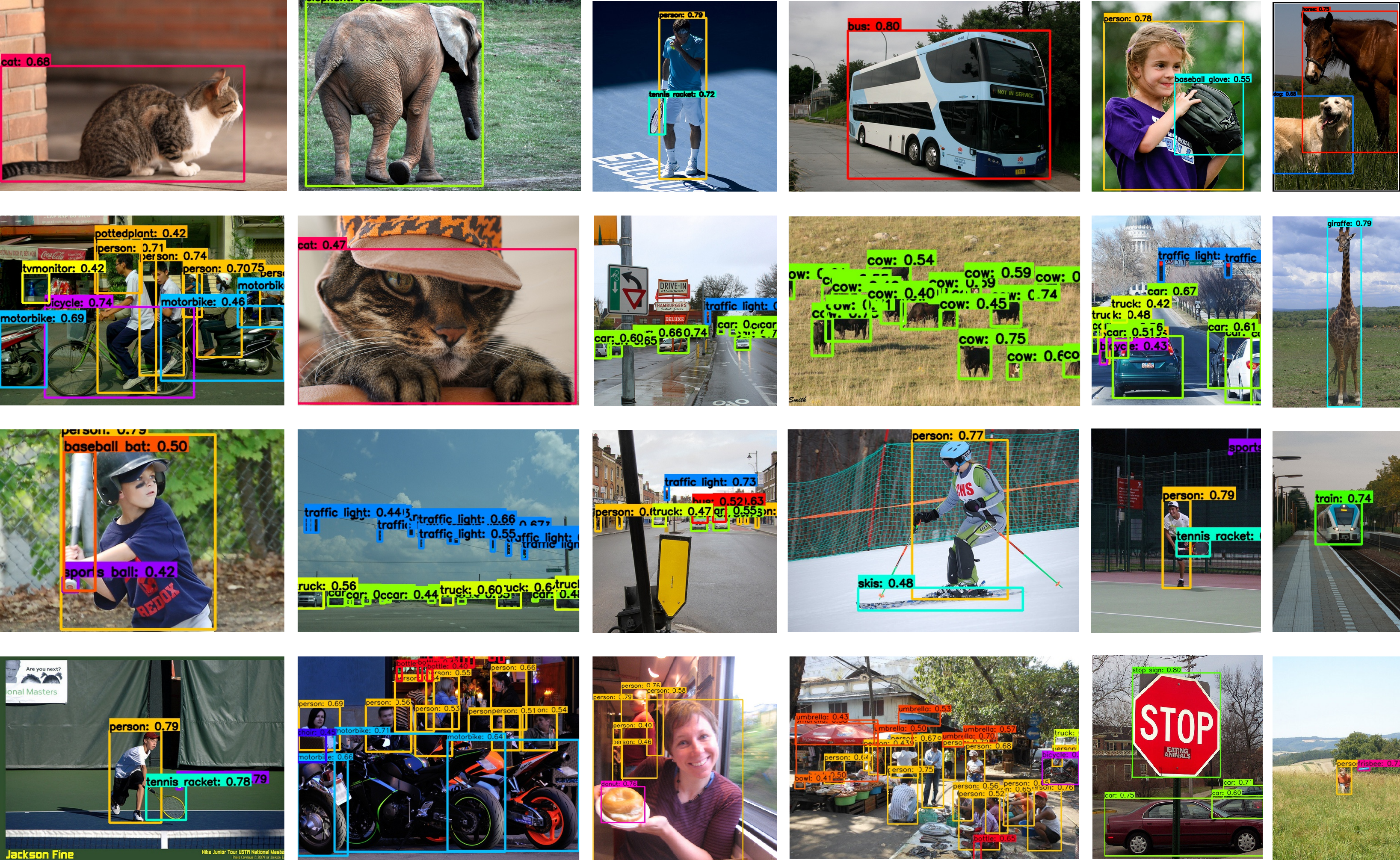} 
        \caption{Quantitive results on the val2017 split. }
    \label{imgdemox} 
    \end{figure*}

Figure \ref{imgdemox} shows the refined detection results of our proposed method (i.e., the model with AP 42.5 in Table (Sec 4.1: 1) ).  One can see that our SALT-Net works well in various scenes. 

\end{document}